\pdfoutput=1
\documentclass[]{macro/neu_msthesis}
\usepackage{tabularx}
\usepackage{caption}
\usepackage{subcaption}
\usepackage{comment}
\usepackage{float}
\usepackage{lipsum}
\usepackage{color}
\usepackage[style=numeric]{biblatex}
\DeclareSourcemap{
  \maps{
    \map{
      \pertype{article}
      \step[fieldset=language, null]
      \step[fieldset=url, null]
      \step[fieldset=issn, null]
      \step[fieldset=isbn, null]
      \step[fieldset=note, null]
      \step[fieldset=editor, null]
      \step[fieldset=urldate, null]
      \step[fieldset=file, null]
    }
  }
}
\DeclareSourcemap{
  \maps{
    \map{
      \pertype{inproceedings}
      \step[fieldset=language, null]
      \step[fieldset=url, null]
      \step[fieldset=issn, null]
      \step[fieldset=isbn, null]
      \step[fieldset=note, null]
      \step[fieldset=editor, null]
      \step[fieldset=urldate, null]
      \step[fieldset=file, null]
    }
  }
}
\DeclareSourcemap{
  \maps{
    \map{
      \pertype{incollection}
      \step[fieldset=language, null]
      \step[fieldset=url, null]
      \step[fieldset=issn, null]
      \step[fieldset=isbn, null]
      \step[fieldset=note, null]
      \step[fieldset=editor, null]
      \step[fieldset=urldate, null]
      \step[fieldset=file, null]
    }
  }
}
\usepackage[usenames,dvipsnames]{xcolor}
\usepackage{tcolorbox}
\usepackage{array}
\usepackage{colortbl}
\usepackage{booktabs}
\usepackage{rotating}

\title{Loco-Manipulation with Non-impulsive Contact-Implicit Planning in a Slithering Robot}

\author{Kruthika Gangaraju}

\nuid{002193742}

\dept{Electrical and Computer Engineering}

\degreename{Robotics - Concentration in Mechanical}

\field{Robotics - Concentration in Mechanical}

\submitdate{April 2024}

\numberofmembers{6}
\principaladviser{Dr. Alireza Ramezani}
\secondadviser{Dr. Gunar Schirner}
\firstreader{Dr. Rifat Sipahi}
\chairman{Dr. Marilyn Minus}
\dean{Dr. Waleed Meleis}

\usepackage{amsfonts,amssymb,amsmath}			
\usepackage{times}		
\usepackage{bm}

\newcommand{\ifno}[1]{}

\usepackage{multirow}
\makeindex
\usepackage{makeidx}
\usepackage{acronym}

\usepackage{url}
\usepackage{epstopdf}
\usepackage[]{graphicx}

\ifnum\pdfoutput>0
\usepackage[pdftex,
bookmarks=true,
bookmarksnumbered=true,
hypertexnames=false,
breaklinks=true           
]{hyperref}
\else
\usepackage[hypertex,
bookmarks=true,
bookmarksnumbered=true,
hypertexnames=false,
breaklinks=true           
]{hyperref}
\fi

\hypersetup{				
pdfauthor = {\authorRef},
pdftitle = {\titleRef},
pdfsubject = {\expandafter{\degreeRef} thesis submitted to Northeastern University},
pdfkeywords = {add keywords here}
pdfcreator = {LaTeX with hyperref package},
pdfproducer = {dvips + ps2pdf}}

\usepackage{microtype}

\begin{document}

\pdfbookmark[1]{Cover}{cover}

\titlepage

\begin{frontmatter}


\begin{dedication}
To Mom, Dad and Tanmay.
\end{dedication}


\pdfbookmark[1]{Table of Contents}{contents}
\tableofcontents
\listoffigures
\newpage\ssp


\chapter*{List of Acronyms}
\addcontentsline{toc}{chapter}{List of Acronyms}

\begin{acronym}
\acro{DoF}{Degrees of Freedom}.

\acro{CPG}{Central Pattern Generator}.

\acro{ODE}{Ordinary Differential Equation}.

\acro{IMU}{Inertial Measurement Unit}.

\end{acronym}


\begin{acknowledgements}

This work has been supported by a lot of people in a number of different ways. I would first like to acknowledge the mentorship and guidance provided to me by my advisor Dr. Alireza Ramezani over the course of this thesis, opening up new opportunities and perspectives for research. I would also like to acknowledge my co-advisors Dr. Gunar Schirner and Dr. Rifat Sipahi for their guidance. I would like to thank my mentor, Adarsh Salagame, for his constant support and encouragement that made the completion of this thesis possible. I would also like to acknowledge the support of other Silicon Synapse lab members, Harin Nallaguntla, Aditya Bondada, Shreyansh Pitroda, Bibek Gupta, Yogi Shah, Henry Noyes, Nolan Smithwick, Alex Qui and few others who helped with simulation and experimentation. I also want to thank my friends Komal Sajwan, Jhanvi Chande, Devansh Dedhia, Dheer Kachalia, Shreya Rahate, Aditya Gosalia, Pratyush Padhi, Ronak Bhanushali and Prem Sukhadwala for their constant support and motivation, and most importantly believing in me. Finally I would like to thank my parents, without whom none of this would be possible, my family for their encouragements and especially my brother Tanmay, who has been a constant pillar of support for me.

\end{acknowledgements}


\begin{abstract}
{
Object manipulation has been extensively studied in the context of fixed base and mobile manipulators. However, the overactuated locomotion modality employed by snake robots allows for a unique blend of object manipulation through locomotion, referred to as loco-manipulation. The following work presents an optimization approach to solving the loco-manipulation problem based on non-impulsive implicit contact path planning for our snake robot COBRA. This thesis presents the mathematical framework and show high-fidelity simulation results and experiments to demonstrate the effectiveness of our approach.
}

\end{abstract}

\end{frontmatter}

\pagestyle{headings}


\chapter{Introduction}
\label{chap:intro}
Object manipulation is an overlap of mechanical design, control engineering and sensor fusion aimed at providing robots with human-like ability to interact with the world. It encompasses a broad spectrum of tasks, ranging from simple pick-and-place operations to complex manipulation and assembly processes in industrial settings, as well as more sophisticated interactions in human-robot collaboration scenarios and domestic environments.
Object manipulation in robots can be achieved with the use of different types of end effectors to the manipulators to achieve desired tasks. These can be in the form of claws, suction cups, or finger-like attachments.

\begin{figure}
    \centering
    \includegraphics[width=1.0 \linewidth]{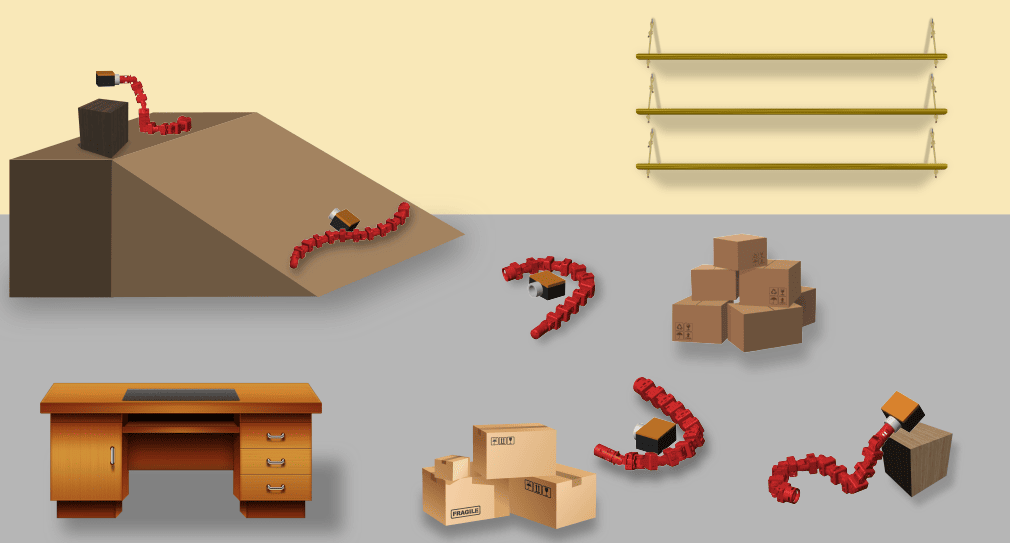}
    \caption{Cartoon illustration of loco-manipulation problem.}
    \label{fig:loco-man}
\end{figure}

In addition to hardware, object manipulation in robotics relies heavily on software algorithms and control systems to plan and execute manipulation tasks effectively. Machine learning and computer vision techniques play a crucial role in enabling robots to perceive and understand the environment, recognize objects, and plan appropriate manipulation strategies \cite{elguea-aguinaco_review_2023}.
One of the key challenges in robotic object manipulation is achieving robustness and adaptability across diverse environments and objects. Robots must be able to handle variations in object size, shape, weight, and texture, as well as unforeseen obstacles and disturbances in the environment. This requires the development of algorithms for adaptive grasping and manipulation, as well as robust sensing and feedback mechanisms to detect and correct errors in real-time.
Object manipulation has a wide range of applications across various sectors, where the ability to grasp, move, and interact with objects is essential. Some of the sectors where object manipulation is particularly useful include:
\begin{enumerate}
    \item Manufacturing and Industrial Automation: Object manipulation is fundamental in manufacturing for tasks such as assembly, pick-and-place operations, packaging, and material handling. Robots equipped with manipulation capabilities increase efficiency, consistency, and throughput in production lines while reducing labor costs and minimizing errors \cite{arents_smart_2022}.
    \item Healthcare and Medical Robotics: Object manipulation plays a vital role in healthcare for tasks such as surgical procedures, patient care, and laboratory automation. Surgical robots assist surgeons in performing minimally invasive surgeries with greater precision and dexterity, while robotic systems handle tasks such as medication dispensing, sample handling, and sterilization \cite{holland_service_2021}.
\end{enumerate}

\subsection{Challenges of Object Manipulation}
Object manipulation in robotics presents several challenges, stemming from the complexity of interacting with the physical world in dynamic and uncertain environments \cite{billard_trends_2019}. Some of the key challenges include:
\begin{enumerate}
    \item Variability in Object Properties: Objects encountered in real-world environments exhibit variability in terms of size, shape, weight, texture, and material properties. Robots must be able to adapt their manipulation strategies to handle this variability effectively. Developing algorithms that can generalize across different object types and properties remains a significant challenge. 
    \item Grasping and Manipulation Planning: Planning and executing grasping and manipulation actions in real-time require solving complex optimization problems. Robots must determine suitable grasp points and orientations for objects, considering factors such as stability, accessibility, and task constraints. Developing efficient algorithms for grasp planning and manipulation sequencing remains an ongoing research area. 
    \item Adaptation to Uncertainty: Real-world environments are characterized by uncertainty, including variations in object pose, unexpected obstacles, and disturbances. Robots must be able to adapt their manipulation strategies in response to uncertainty to achieve robust and reliable performance. This may involve incorporating feedback mechanisms or employing predictive models to anticipate and mitigate uncertainty. 
    \item Dexterity and Manipulation Skills: Achieving dexterous manipulation requires precise control over the robot's end-effector, whether it's a robotic hand, gripper, or manipulator. Designing robotic hands capable of emulating human-like dexterity while ensuring mechanical robustness and efficiency remains a significant engineering challenge. Moreover, developing control algorithms that enable fine-grained manipulation and coordination of multiple degrees of freedom is non-trivial. 
    \item Learning and Adaptation: Object manipulation tasks may involve handling novel objects or operating in unfamiliar environments. Robots must be able to learn from experience and adapt their manipulation strategies over time to improve performance and adapt to new situations. Developing efficient and scalable learning algorithms that can leverage large datasets or simulation environments while ensuring safety and reliability is a significant research challenge. 
\end{enumerate}

\section{Loco-manipulation}
"Locomotion + manipulation" refers to the integration of both mobility and manipulation capabilities in robots. It combines the ability to move through an environment (locomotion) with the capability to interact with objects within that environment (manipulation). This integration enables robots to perform a wide range of tasks that involve both navigating through space and interacting with objects.

In many robotic systems, locomotion and manipulation are treated as separate subsystems, each with its own set of sensors, actuators, and control algorithms. Locomotion systems enable robots to move from one location to another, whether it's navigating across rough terrain, climbing stairs, or traversing indoor environments. These locomotion mechanisms can range from wheeled or tracked platforms to legged robots or even flying drones.

On the other hand, manipulation systems allow robots to grasp, lift, move, and manipulate objects in their environment. This may involve using robotic arms with grippers, hands, or specialized end-effectors to interact with objects in various ways. Manipulation capabilities are essential for tasks such as pick-and-place operations, assembly, sorting, and handling objects in unstructured environments.

The integration of locomotion and manipulation capabilities enables robots to perform complex tasks that require both mobility and manipulation skills. For example:

\textbf{Search and Rescue:} Robots equipped with both mobility and manipulation capabilities can navigate through disaster zones or hazardous environments to search for survivors while also manipulating debris or obstacles to access hard-to-reach areas \cite{han_snake_2022}.

\begin{figure}
    \centering
    \includegraphics[width=0.6 \linewidth]{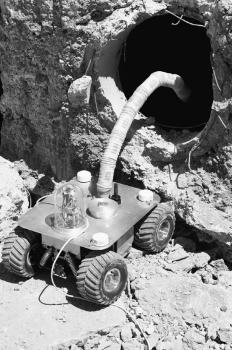}
    \caption{Illustrates a hyper-redundant robot mounted on a mobile base from A. Wolf, et al., 2005 \cite{wolf_design_2005}, in a search and rescue mission.}
    \label{fig:sar}
\end{figure}

\textbf{Warehouse Automation:} Robots in warehouse environments can navigate aisles to locate items, pick them up, and transport them to the desired location for sorting, packing, or shipping \cite{duz_robotic_2022}.

\begin{figure}
    \centering
    \includegraphics[width=0.6 \linewidth]{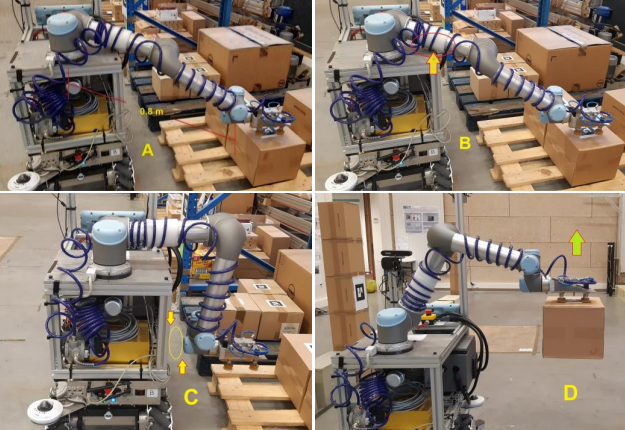}
    \caption{Illustrates robot in a warehouse, M. A. Khan, 2020, \cite{khan_design_nodate}}
    \label{fig:warehouse}
\end{figure}

\textbf{Construction and Maintenance in space:} Construction and maintenance tasks in space are critical for the establishment and upkeep of infrastructure essential for space exploration and habitation. In these endeavors, robots play a pivotal role in executing various functions necessary for building, repairing, and maintaining space structures and facilities. These tasks include transporting construction materials, assembling intricate components, and conducting routine maintenance operations. Robots navigate through the unique challenges of the space environment, utilizing advanced technologies to maneuver, interact with, and manipulate tools and equipment. Through their capabilities, robots contribute to the ongoing development and sustainability of human presence and activities in space \cite{papadopoulos_robotic_2021}.

\begin{figure}
    \centering
    \includegraphics[width=0.6 \linewidth]{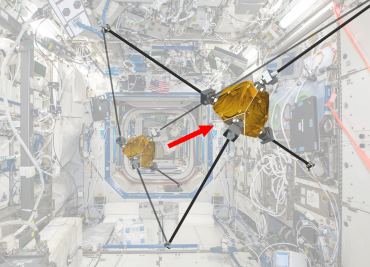}
    \caption{Illustrates robot navigating around space station, S. Schneider, et al., 2021, \cite{schneider_reachbot_2021}}
    \label{fig:construction}
\end{figure}

\textbf{Assistive Robotics:} In assistive applications, robots can provide mobility assistance to individuals with disabilities while also offering manipulation capabilities to help with tasks such as picking up objects, opening doors, or operating household appliances \cite{chen_robots_nodate}.

Integrating locomotion and manipulation capabilities in robots poses several technical challenges, including coordinating motion and manipulation actions, maintaining balance and stability during manipulation tasks, and adapting to changes in the environment. However, advancements in robotics technology, including improvements in sensors, actuators, and control algorithms, continue to enable the development of increasingly capable robots that can effectively combine locomotion and manipulation to perform complex tasks in diverse real-world scenarios.
\subsection{Challenges in locomanipulation}
\begin{figure}
    \centering
    \includegraphics[width=1.0 \linewidth]{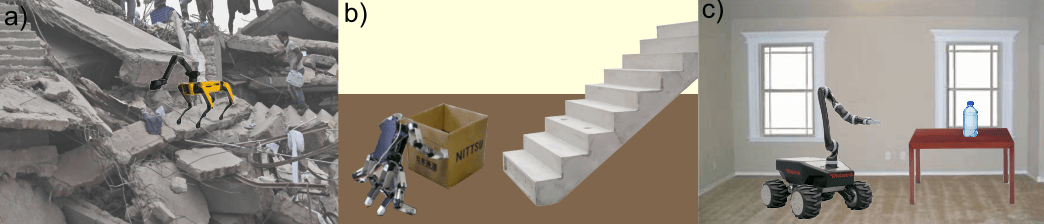}
    \caption{Challenges of loco - manipulation; a) shows the difficulty of navigating through clustered, narrow spaces \cite{noauthor_boston_nodate}; b) shows the challenge of taking a box up a flight of stairs \cite{inoue_pushing_2010}; c) shows a wheeled manipulator trying to pick up an object during its locomotion \cite{noauthor_xl-gen_nodate}.}
    \label{fig:challenges}
\end{figure}
Integrating locomotion and manipulation capabilities in robots presents several challenges, spanning technical, computational, and practical aspects. Some of the key challenges include:
\begin{enumerate}
    \item \textbf{Coordination of Locomotion and Manipulation:} A major problem for solving locomotion challenge is to ensure stability of the robot during its interactions with environment. For instance, a legged robot has to make sure its projected center of mass should fall into the support polygon. On top of it, when combining manipulation tasks, the weight of the object has to be taken into consideration and made sure that the entire weight of the robot during this operation should be inside the support polygon to ensure stability. Also the contacts between the manipulator and the object should be well defined to ensure sufficient grasping even during locomotion of the mounted manipulator \cite{gong_legged_2023}.
    \item \textbf{Sensor Integration and Fusion:} Integrating locomotion and manipulation requires addition of multiple sensors in order to achieve efficiency. Sensors like Ultrasonic range findes, lidars can be used for estimating the distance from object, identifying obstacles; stereo cameras or 3D cameras can be integrated for localization of the mobile robot, integration of force and tactile sensors on manipulator joints to perfectly estimate the amount of force required to be applied to achieve necessary grasping. The challenge of having this type of multi-sensor fusion is due to the extremely dynamic nature of these mobile manipulators, there could be noise in data which gives rise to inaccurate measurements and error in precision during operation \cite{zhou_learning-based_2022}.
    \item \textbf{Path Planning, Trajectory Optimization and Manipulation in Unstructured Environments:} Planning optimal paths for both locomotion and manipulation actions in complex environments involves solving high-dimensional planning and optimization problems. Generating collision-free paths while considering constraints such as kinematic limits, environmental obstacles, and task requirements requires efficient algorithms capable of handling real-time planning and replanning. Manipulating objects in unstructured or cluttered environments, such as construction sites or disaster zones, presents additional challenges. Robots must be able to adapt their manipulation strategies to handle variations in object shape, size, weight, and location, as well as navigate through obstacles to reach target objects. 
    \item \textbf{Physical Interaction and Force Control:} Ensuring safe and effective physical interaction with objects during manipulation tasks requires precise force and torque control. Robots must be able to apply the right amount of force and pressure to grasp, lift, or manipulate objects without causing damage or instability. Developing robust control algorithms for force sensing and feedback control is essential for successful manipulation in diverse scenarios. 
    \item \textbf{Energy Efficiency and Payload Capacity:} Integrating locomotion and manipulation capabilities while maintaining energy efficiency and payload capacity is a significant challenge, especially for mobile robots or drones with limited onboard power and weight constraints. Optimizing the design and control of the robot's actuators, mechanisms, and power systems to minimize energy consumption while maximizing payload capacity is crucial for practical deployment in real-world applications. 
\end{enumerate}

\section{Classification of Loco-Manipulation Approaches}

\begin{figure}
    \centering
    \includegraphics[width=0.6\linewidth]{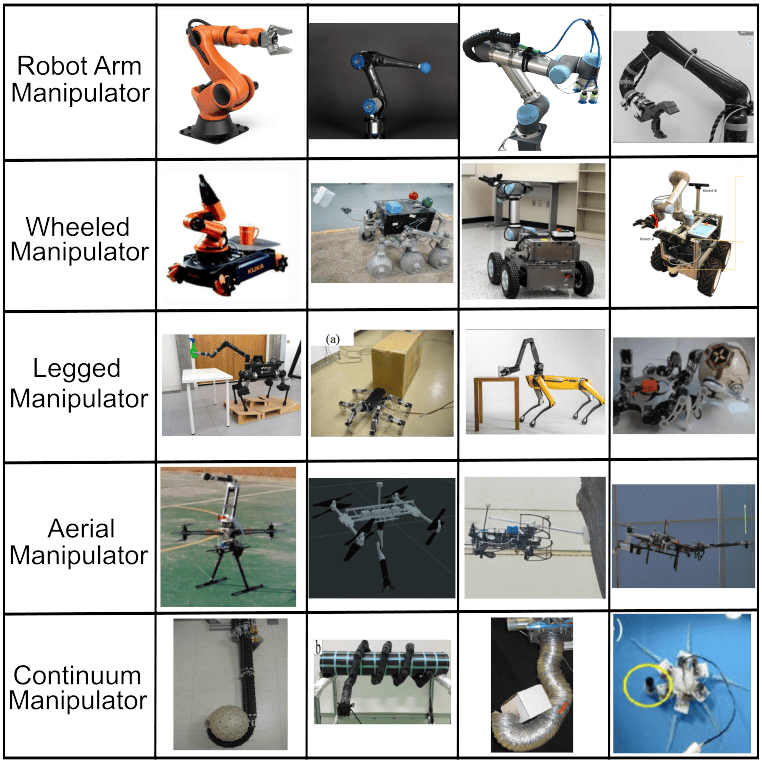}
    \caption{Classification of object manipulation approaches in the literature}
    \label{fig:classification}
\end{figure}
\subsection{Robot Arm Manipulators}
There have been different manipulators in development over decades. Unimate introduced the first industrial robotic arm in 1961, it has subsequently evolved into the PUMA arm. This gave rise to further developments in robotic arms over the years \cite{moran_evolution_2007}. Recent developments in these manipulators include having modifications in modelling to allow more flexibility and workspace. There are also various algorithms developed based on the environments in which these operate \cite{satya_durga_manohar_sahu_modelling_2022, bodie_anypulator_2016} . Another development to these robotic arm manipulators is to have a moving base which increases the workspace \cite{kumar_linear_2023, chand_vertically_2022}. While these manipulators excel in tasks such as "pick and place" operations within controlled environments, they often fall short when extensive motion range is required. This is where loco-manipulation proves invaluable. 
\subsection{Wheeled Mobile Manipulators}
There are various methods of combining the task of manipulation to a mobile robot. Firstly we explore the straightforward way of achieving loco-manipulation - having a manipulator on a wheeled system \cite{thakar_survey_2023} \cite{ben-tzvi_design_2008} . These systems provide sufficient stability during movement and a wide range of motion. Recent advancements also enable manipulation whilst moving. Despite its high feasibility, wheeled systems fail in highly unstructured environments and constrained spaces. They also introduce the problem of altering the environment due to their wheels which may not be desirable in all cases.  This leads to exploring another method of incorporating manipulation on locomotors i.e. with the use of legged robots. 
\subsection{Legged Mobile Manipulators}
Legged robots can have a unique role in manipulating objects in dynamic, human-centric, or otherwise inaccessible environments \cite{gong_legged_2023, lopes_review_2023}. They are more efficient in dodging obstacles as they can adjust their height and have fewer floor contact areas. There are multiple ways to use a legged system for object manipulation - 1. Passive manipulation i.e. pushing or kicking the object to desired location \cite{inoue_pushing_2010} \cite{flemmer_humanoid_2014}, 2. Grasping with legs \cite{heppner_laurope_nodate} \cite{hirose_quadruped_2005} \cite{shaw_keyframe-based_2022}, 3. Attaching manipulators to the legged system, this is a more commonly used approach, for example Spotmini \cite{gueguen_hey_2016} and ANYmal \cite{ferrolho_optimizing_2020} have been equipped with manipulators on their bodies to perform loco-manipulation tasks. One interesting robot is the TerminatorBot \cite{voyles_terminatorbot_2005} which was specifically designed for the purpose of combining locomotion with manipulation. However these systems fail due to instability problems that may arise. If the legged system is designed to transform from purely legged configuration to manipulator + legged configuration, decreased contacts may make it unstable whereas if there is a separate manipulator attached to the legged system, there is a limit to the weight the manipulator can lift for it to maintain the entire center of mass within the support polygon to maintain stability. 

\begin{figure}
    \centering
    \includegraphics[width=0.6 \linewidth]{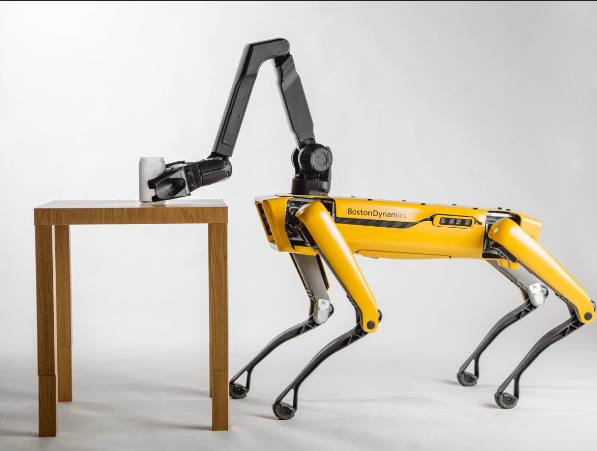}
    \caption{Illustrates Spot picking object from table \cite{gong_legged_2023}}
    \label{fig:loco-man-legged}
\end{figure}

\subsection{Aerial Manipulators}
To entirely eliminate these problems of having contacts with the environments one way to approach this problem is to have an aerial manipulator \cite{jimenez-cano_aerial_2015, kim_globally_2023}. Aerial manipulators encounter analogous challenges to those of UAVs, including nonlinearities, strong coupling, under-actuation, and susceptibility to disturbances \cite{meng_survey_2020}.
In order to eliminate the above mentioned problems and increase versatility, continuum manipulators came into picture. These manipulators have a wide range of motion and due their flexible nature, they are not limited to operating conditions. 

\begin{figure}
    \centering
    \includegraphics[width=0.6 \linewidth]{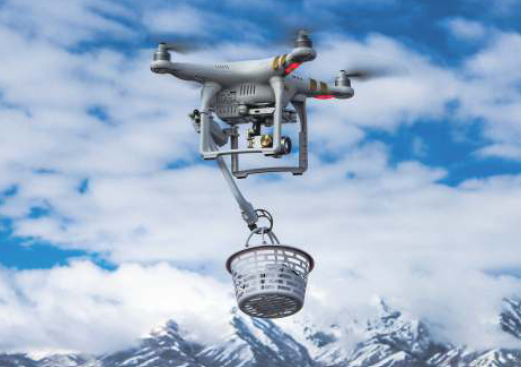}
    \caption{Illustrates a drone carrying packages \cite{ramalepa_review_2021}}
    \label{fig:loco-man-aerial}
\end{figure}

\subsection{Continuum Manipulators}
Some research has been done on incorporating continuum manipulators on locomotive systems \cite{kumar_singh_continuum_2014} \cite{giri_continuum_2011} \cite{hannan_elephant_2001} \cite{hirose_snake-like_2009}. Using a snake robot as a manipulator becomes an interesting task since due to its high dimensionality and slender structure, it can operate in any type of unstructured environment. \cite{yamakita_control_nodate} \cite{elsayed_mobile_2022} \cite{wang_unified_2011}. Also the different gaits that can be implmented using snake-like locomotion on these robots gives flexibility in choosing the most energy efficient method in which the object can be manipulated \cite{wiriyacharoensunthorn_analysis_2002, ma_analysis_2003, ohno_design_2001, liljeback_modular_2005, burdick_sidewinding_1994, shan_design_1993, yim_new_1994, rincon_ver-vite_2003}.  However, looking at the challenges faced by commonly used mobile manipulators and the problem we are trying to solve, using a snake robot for the purpose of object manipulation is the area I decided to explore.

\section{Prior SS Lab Works}
Before embarking on the creation of a versatile system adept at handling various locomotion and manipulation tasks, I drew inspiration from the diverse array of projects underway at Northeastern Silicon Synapse Lab. Engaging in control work on these projects equipped me with insights to tackle the locomotion and manipulation challenges, taking inspiration from biological systems \cite{salagame_letter_2022, salagame_quadrupedal_2023, ramezani_atrias_2012, noauthor_mabel_2011,ramezani_bat_2016, sihite_mechanism_2020, ramezani_biomimetic_2017}. At Northeastern, projects like Husky Carbon, a morphofunctional quadruped utilizing thrusters for stability in demanding environments \cite{sihite_dynamic_2023, sihite_efficient_2022, sihite_optimization-free_2021, dangol_thruster-assisted_2020, }, and Harpy, a biped employing thruster-assisted walking, provided valuable reference points  \cite{sihite_unilateral_2021, dangol_control_2021, de_oliveira_thruster-assisted_2020,  noauthor_feedback_nodate, dangol_thruster-assisted_2020, park_finite-state_2013}. Additionally, collaborative efforts with Caltech led to the development of the Multimodal Mobility Morphobot (M4), capable of seamlessly transitioning between aerial and wheeled locomotion \cite{sihite_efficient_2022, sihite_multi-modal_2023, sihite_demonstrating_2023, liang_rough-terrain_2021}. Northeastern's Aerobat showcased proficiency in navigating confined spaces where traditional drones struggle due to rotor-induced backdraft \cite{sihite_wake-based_2022, sihite_enforcing_2020, sihite_bang-bang_2022,  syed_rousettus_2017, sihite_orientation_2021, sihite_computational_2020, sihite_unsteady_2022, sihite_integrated_2021, }. Across all these platforms, robust control strategies are imperative, facilitating effective planning and seamless morphing between different locomotive configurations. \nocite{ramezani_describing_2017, ramezani_generative_2021, ramezani_lagrangian_2015, ramezani_modeling_2016, ramezani_nonlinear_nodate, ramezani_performance_2014, ramezani_towards_2020, sihite_actuation_2023, sihite_bang-bang_2022, sihite_computational_2020, sihite_demonstrating_2023, sihite_dynamic_2023, sihite_efficient_2022, sihite_enforcing_2020, sihite_integrated_2021, sihite_mechanism_2020, sihite_multi-modal_2023, sihite_optimization-free_2021, sihite_orientation_2021, sihite_unilateral_2021, sihite_unsteady_2022, sihite_wake-based_2022}

\section{COBRA Design Motivation}
Snake robots have undergone over three decades of development, largely due to their adaptable nature across various applications. This prompted the creation of COBRA specifically for NASA's BIG Idea competition. In 2022, NASA's Innovative Advanced Concepts (NIAC) Program invited academic institutions in the United States to participate in NASA's Breakthrough, Innovative, and Game-changing (BIG) Idea competition, with a focus on pioneering mobility solutions capable of traversing lunar craters. Northeastern University clinched NASA's prestigious Artemis Award in this competition with a groundbreaking proposal introducing an articulated robot tumbler named COBRA (Crater Observing Bio-inspired Rolling Articulator). COBRA employs sidewinding motion to approach the crater, transforms into a wheel configuration, and utilizes gravitational forces to descend into the crater, thereby establishing an energy-efficient system while ensuring stability. These distinctive attributes of COBRA render it applicable across diverse domains, including search and rescue operations, inspection within confined spaces, and the manipulation of objects in cluttered environments.

\section{Loco-manipulation using COBRA}

Although using a snake robot as a manipulator is an explored avenue, what makes this thesis novel is the design of COBRA itself. Most of the snake robots in existence either have wheels or do not have the capability of grasping onto an object and moving around. The existence of wheels defeats the purpose of operating in unstructured environments. The absence of wheels on COBRA and the ability for it to latch onto objects makes it easier to develop gaits which can hold on to objects of different shapes and sizes and move those objects around in different environments with rough terrains or inclines.

\begin{figure}
    \centering
    \includegraphics[width=1.0\linewidth]{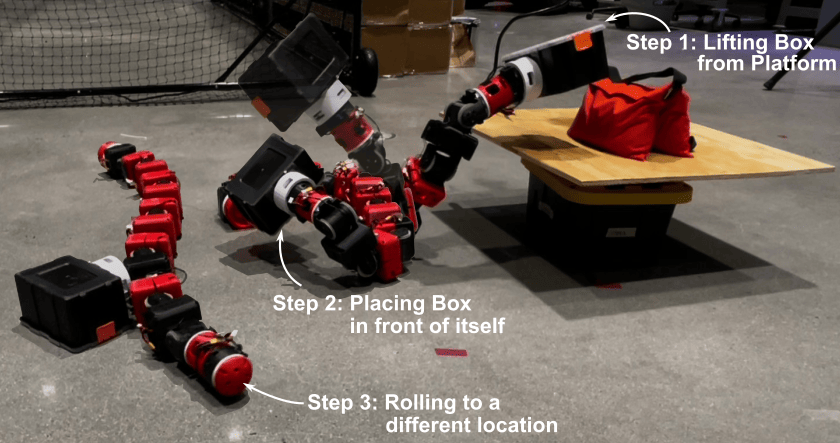}
    \caption{Object manipulation using COBRA}
    \label{fig:COBRA manip}
\end{figure}

\section{Objectives and Outline of Thesis}
\label{sec:objectives}

Optimization-driven path planning and control strategies have emerged as pivotal methodologies for managing diverse contact-intensive systems within real-world experimental settings. These approaches have found widespread application across various locomotion modalities, such as legged and slithering locomotion, showcasing remarkable efficacy, including rapid contact planning in terrestrial environments \cite{ding_real-time_2019,carius_trajectory_2018, droge_optimal_2012}. Notably, point-contact models such as legged robots have been particularly receptive to optimization techniques \cite{sleiman_versatile_2023} \cite{alipour_dynamically_2015} compared to systems characterized by extensive contact interactions, such as snake robots. Given the intricate dynamics inherent in slithering systems, which encompass sophisticated contact dynamics \cite{fu_robotic_2020,chen_studies_2004, sleiman_contact-implicit_2019}, there arises a pressing need for enhanced modeling and control methodologies. These tools are indispensable for orchestrating body movements through the modulation of joint torques, ground reaction forces, and the coordination of contact sequences comprising timing and spatial positioning. Addressing this contact-rich problem presents intriguing prospects for leveraging contact-implicit optimization, which represents a prevalent design paradigm in the field of locomotion and for unknown reasons is less explored in snake-type robots. The primary research objectives encompass: Investigating optimal control design approaches to effectively guide the joints along desired trajectories for object manipulation. This work introduces an optimization approach based on non-impulsive contact-implicit path planning for COBRA. We demonstrate the effectiveness of this method in generating optimal joint trajectories for desired object movements across flat and ramp surfaces in simulation and experiment \cite{pozzi_modeling_2022, sahu_gait_2019, noauthor_snake_nodate, sibilska-mroziewicz_framework_2022}.


\section{Contributions}
\label{sec:contributions}
In this thesis, I demonstrated the application of COBRA within the Simscape model to manipulate objects through various methods. The research showcases the manipulation of boxes with differing weights, both on level ground and a sloped ramp. Through meticulous experimentation, it is illustrated that COBRA exhibits proficiency in lifting boxes and relocating them onto raised platforms, and conversely, transferring them from elevated positions to ground level. The experiments conducted within the simulation environment were meticulously replicated on physical COBRA hardware, yielding consistent results. This process not only validates the efficacy of COBRA within a virtual setting but also underscores its practical applicability in real-world scenarios. The study underscores the versatility of COBRA in object manipulation tasks, demonstrating its adaptability to diverse environments and scenarios. By showcasing its capabilities across simulated and physical domains, the research bridges the gap between theoretical simulations and practical implementations. Furthermore, the successful replication of results on physical hardware reaffirms the reliability and robustness of COBRA-based manipulation techniques. This not only bolsters confidence in the simulation-based findings but also instills trust in the potential deployment of COBRA in real-world applications requiring precise object manipulation. Overall, my thesis contributes to the understanding of COBRA's capabilities within object manipulation, providing valuable insights for both theoretical modeling and practical implementation. Through meticulous experimentation and validation, it demonstrates the effectiveness of COBRA across simulated and physical environments, paving the way for its utilization in various industrial and robotic applications.


\chapter{COBRA Platform}
\label{chap:cobra}

\section{COBRA Hardware}
\label{section:hardware}
COBRA has 11 degrees of freedom that enables the robot to change into different shapes enabling different modes of locomotion. In snake mode, our system will utilize sidewinding for movement on flat or uphill terrain. Sidewinding is a type of snake locomotion used to move across loose or slippery surfaces such as sand. The symmetry of the modules allows us to make COBRA roll in various directions without having to make any modifications to the module design. It can also enter the tumbling mode by connecting the head and tail of the system together.

\begin{figure}
    \centering
    \includegraphics[width=1\linewidth]{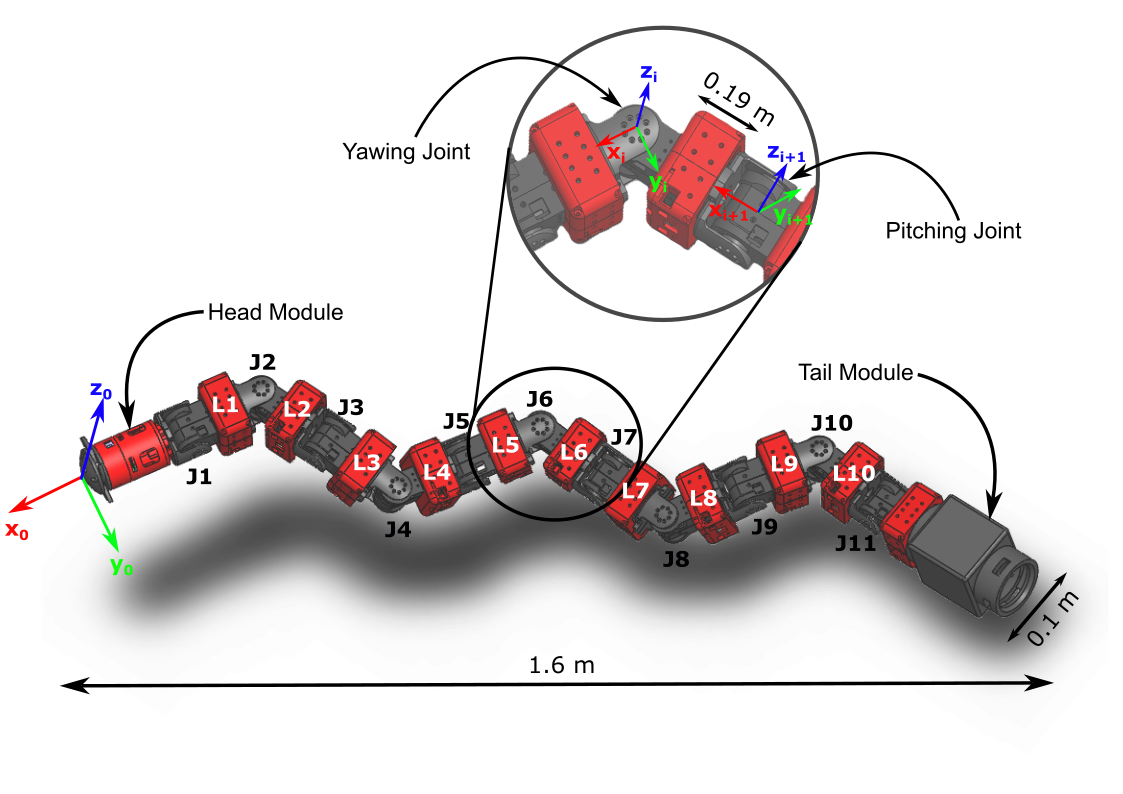}
    \caption{This image shows the overview of COBRA model with the head coordinate frame and the coordinate frames for the yawing and pitching joints which are consistent throughout the robot. It also show the naming convention used accross all the models and codes in which J$x$ corresponds to the joints and L$y$ corresponds to the link where $x$ ranges from 1 to 11 and $y$ ranges from 1 to 10, excluding the head and tail.}
    \label{fig:COBRA modes}
\end{figure}

A modular system provides many benefits for operation, design, and manufacturability. The goal is that each joint module is simple and interchangeable, allowing for rapid prototyping and building of the system. As shown in Fig. \ref{fig:module} below, a single module of COBRA has 1 degree of freedom joint that can rotate a full 180 degrees. In addition, each module has a circular female connector on one end and a male connector on the other. This way the modules can be attached in different configurations. The
system will consist of 10 joint modules alternately connected. The 11th module
is the tail module which is designed to latch onto the head module to morph into the tumbling configuration. The entire system weighs only 6 kg, with a diameter of 10 cm and length of 1.6 meters.

The joint modules utilize XW series motors from DYNAMIXEL. These off-the-shelf servo motors are already integrated with a controller, driver, encoder, reduction gear, and RS-485 network communication. The XW series also have IP68 certified ingress protection from dust and are waterproof to depths of 1 meter. The power and communication for these motors are daisychained and controlled by a single Raspberry Pi computer located in the head module of the system. This makes it quicker to set up compared to a more custom motor solution. The joint module housing is fabricated using a Markforged carbon fiber reinforced nylon, which allows for extremely strong and lightweight modules. The design of the joint modules is driven by the worst-case loading scenario of transforming into and out of tumbling mode, which requires ~6.9 Newton-meters of torque.

\begin{figure}
    \centering
    \includegraphics[width=0.6\linewidth]{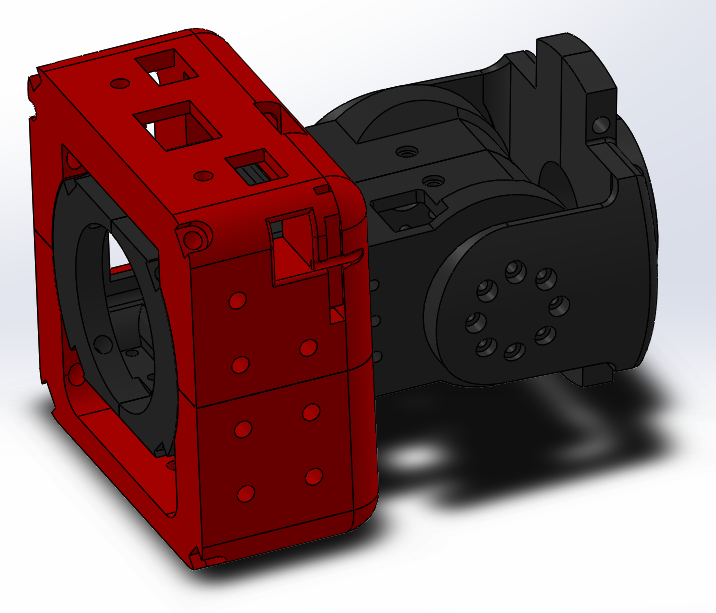}
    \caption{A single COBRA module}
    \label{fig:module}
\end{figure}

\begin{figure}
    \centering
    \includegraphics[width=0.5\linewidth]{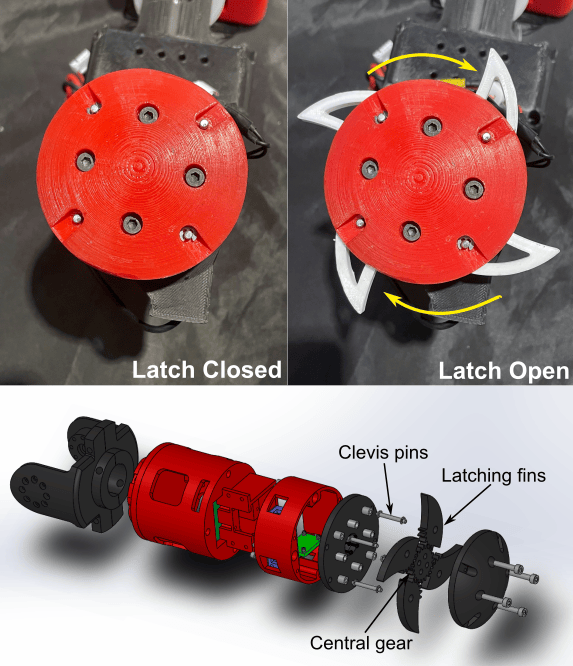}
    \caption{(Above) Closeup view of the head with actuated fins. (Below) Expanded view of the head module.}
    \label{fig:head}
\end{figure}

\begin{figure}
    \centering
    \includegraphics[width=0.3\linewidth]{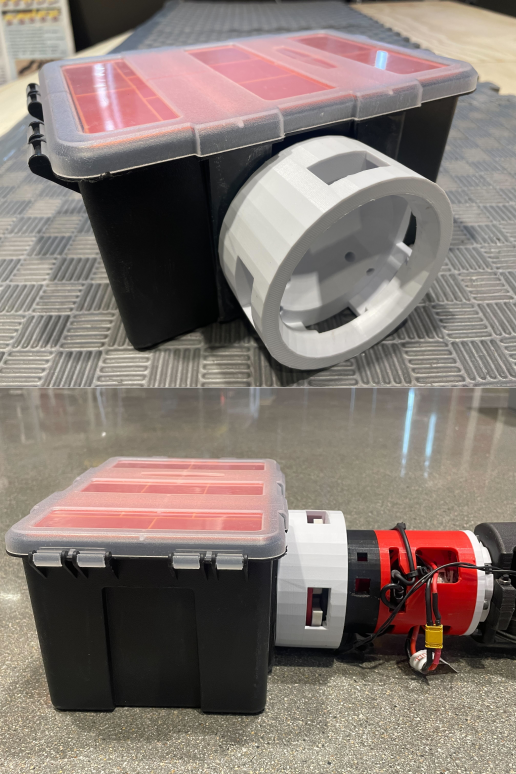}
    \caption{(Above) Docking module on the box. (Below) Head module latched onto the docking module on the box.}
    \label{fig:docking module}
\end{figure}

In addition to the eleven identical modules, COBRA features a distinct module at the snake’s head, aptly referred to as the “head module,” and similarly, a “tail module” at the snake’s tail end. The head module is shown in Fig. \ref{fig:head} . The primary purpose of these unique modules are to connect together to form a loop prior to the onset of tumbling mode. The head module acts as the male connector and utilizes a latching mechanism to sit concentrically inside the female tail module. The latching mechanism consists of a Dynamixel XC330 actuator, which sits within the head module and drives a central gear. This gear interfaces with the partially geared sections of four fin-shaped latching “fins.” The curved outer face of each latching fin has an arc length equal to 1/4 of the circumference of the head module’s circular cross-section. When the mechanism is retracted, these four fins form a thin cylinder that coincides with the head module’s cylindrical face. A dome-shaped cap lies on the end of the head module so that the fins sit between it and the main body of the head module. Clevis pins are used to position the fins in this configuration. COBRA’s tail module features a female cavity for the fins. When transitioning to tumbling mode, the head module is positioned concentrically inside the tail module using the joint’s actuators, and the fins unfold into the cavity to lock the head module in place. For the head and tail modules to unlatch, the central gear rotates in the opposite direction, and the fins retract, allowing the system to return to sidewinding mode. The choice for an active latching mechanism design stemmed from the design requirements and restrictions. Magnets were initially discussed as a passive latching option, however they would not be effective in conjunction with the ferromagnetic regolith. Further, due to the need to stay in a latched configuration even when a large amount of force is applied to the system during tumbling, a passive system was not chosen, for there would be the risk of unlatching during tumbling. An attachment similar to the tail module was made for the box as a docking module \ref{fig:docking module}. This entire mechanism is useful for the loco-manipulation problem which will be discussed further.

\section{Simulation Setup}
\label{section:simulation}
Multibody simulation plays a pivotal role in addressing kinematic and dynamic challenges. Its application spans across a wide spectrum, encompassing everything from individual mechanical components, to complex mechanisms, and even robotic systems. The multibody model of a mechanical system is structured around rigid and deformable bodies, interconnected via kinematic pairs. These bodies are capable of experiencing substantial translational and rotational displacements. The primary parameters utilized for characterizing bodies within the multibody simulator encompass: a local reference frame with its origin situated at the body's center of mass and aligned with its principal axes, the mass, the inertia tensor within the local reference frame, and additional auxiliary references for delineating constraints, represented through interrelations among motion parameters. Both rigid and flexible bodies are depicted using singular or combinations of multiple blocks to portray their mechanical characteristics. The interconnection of all bodies occurs through joints or appropriate constraints, creating an assembly of an articulated mechanism. The resulting degrees of freedom stem from the kinematic relationships established by the constraint blocks. There are several tools available to create a simulation model. We are using the MATLAB \textit{Simulink Multibody Toolbox} to create a high fidelity simulation for COBRA simulation model.
\subsection{Actuator model}
\label{subsection:actuator}
Each link is represented as a rigid body with a mass of $0.5$ kg, and the inertia matrices are automatically computed by MATLAB based on the geometry, assuming a uniform mass distribution. These geometries are obtained as meshes imported from the SolidWorks model for each link. We use the convex hull of these imported geometries which facilitates modeling contacts between surfaces. The inertia and mass information of these geometries are directly imported from the mesh files. The inertia tensor for each of the ten identical body links is as follows:  ($\mathbf{I}_{xx} = 7.167\times 10^{-4},~\mathbf{I}_{yy} = 8.704\times 10^{-4},~\mathbf{I}_{zz} = 8.626\times 10^{-4}~ \text{kgm}^2$), and the inertia tensors for the head and tail modules are ($\mathbf{I}_{xx} - 4.4562\times 10^{-4},~\mathbf{I}_{yy} = 1.710\times 10^{-3},~\mathbf{I}_{zz} = 1.793\times 10^{-3}~ \text{kgm}^2$) and ($\mathbf{I}_{xx} = 8.182\times 10^{-4},~\mathbf{I}_{yy} = 1.141\times 10^{-3},~\mathbf{I}_{zz} = 1.109\times 10^{-3}~ \text{kgm}^2$) along the primary axes and the mass of each module is taken to be roughly $0.5~ g$. Each link is connected via position-controlled revolute joints, with the axes of these joints defined using the Denavit-Hartenberg parameters. The joint restricts the movement of two arbitrary frames linked to the base and follower frames, allowing only pure rotation along a shared axis. This axis of rotation coincides with the z-axis of the joint's base frame. Both the base and follower frames share the same origin and z-axis. The follower frame rotates around the z-axis. The revolute joint block facilitates the replication of actual motor parameters on the robot by adjusting internal mechanical parameters and setting actuation limits to restrict joint movement. In our model, we have specified actuation upper and lower bounds with limiting position values ranging from $-90^{~o}$ to $90^{~o}$, spring stiffness value to $1 \times 10^{4}~Nm/deg$ and damping coefficient to $10~Nms/deg$. Additionally, the revolute joint block enables us to activate sensing from the joints, allowing measurement of joint angles, velocities, accelerations, and torques. This capability enables us to compare the results obtained from our Simscape simulation with the values obtained from our robot actuators. This setup gives us the COBRA simscape model which can be used for various experiments. Along with this, we designed a latch joint using the weld joint which enables engaging and disengaging a joint while the simulation is running. We have a 6 DOF joint modelling between the head frame and the ground frame which enables motion of COBRA with respect to the ground. We can extract the position data from the frames between each link by connecting it to the transform sensor.  This transform sensor is connected between the frame of each link and the ground frame. The measurement frame option on the sensor allows us to opt any frame with reference to which we want to obtain the measured data in. The selection of the measurement frame does not affect the rotational quantities. The Transform Sensor block has five different options for the Measurement Frame parameter. We selected the measurement frame to be the base frame which is the ground frame in our application. We enabled sensing for the transform sensor to collect data for rotational transform, positions velocities, accelerations, angular velocities and angular accelerations in xyz directions. We have this entire actuator model setup as a library block which allows us to setup various simscape models without having to develop the same actuator repeatedly. 
\begin{figure}
    \centering
    \includegraphics[width=0.8\linewidth]{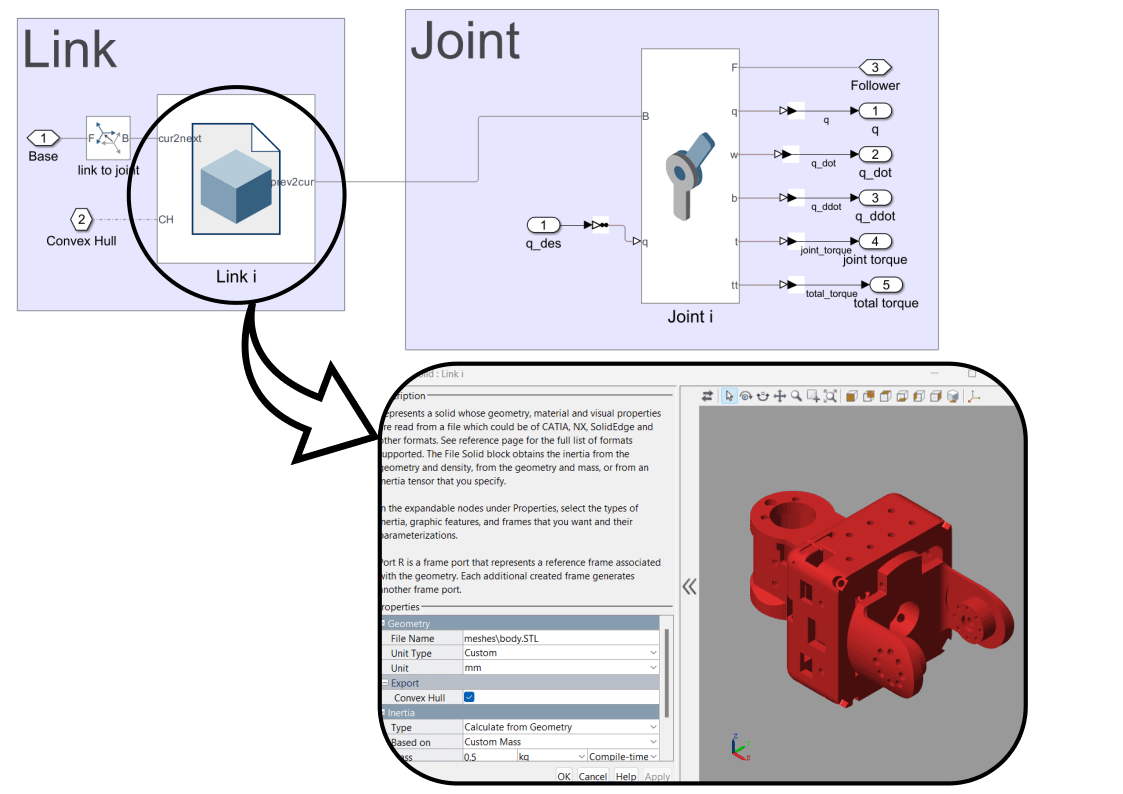}
    \caption{Simscape block for actuator model}
    \label{fig:simscape-actuator}
\end{figure}
\subsection{Contact force model}
\label{subsection:contact force}
The interactions between surfaces of solids is an important aspect in any Multibody simulator. This can be done in Simscape using the Simscape spatial contact force block. This block allows to develop a contact model between a pair of bodies. The Spatial Contact Force block models the contact between a pair of geometries in 3-D space. It uses the Kelvin-Voigt model to model contacts. The model adopts a regularization method, alternatively referred to as compliance or viscoelasticity, wherein the interacting bodies are treated as deformable within the contact region. Consequently, the contact forces can be described as a continuous function of the localized deformation occurring between the contacting surfaces \cite{pozzi_modeling_2022} \cite{skrinjar_review_2018}. Throughout contact, each geometry possesses a dedicated contact frame situated precisely at the contact point. These contact frames consistently coincide and define the contact plane through their xy-planes. Notably, the z-direction of these frames serves as an outward normal vector for the base geometry and an inward normal vector for the follower geometry. As continuous contact ensues, the contact frames dynamically shift along the geometry, tracking the movement of the contact point. The block applies contact forces to the geometries at the origin of the contact frames in accordance with Newton's Third Law: The normal force, $f_n$, which is aligned with the z-axis of the contact frame. This force pushes the geometries apart in order to reduce penetration; The frictional force, $f_f$, which lies in the contact plane. This force opposes the relative tangential velocities between the geometries. The normal forces for all contact interactions between robot links, ground surface and object are modeled using a \textit{Smooth Spring-Damper} model with spring stiffness $1\times 10^{-4}~N/m$, damping coefficient $1\times10^3~Ns/m$ and a transition width of $1\times10^{-3}~m$. Transition width characterises the transitional region to force equations. By varying the transition width we can achieve sharper transitions for lower values and smoother for higher values. The normal force is calculated by: 
$$
f_{n}=s ( d, w ) \cdot( k \cdot d+b \cdot d^{\prime} )
$$
where,
\begin{itemize}
\addtolength\itemsep{-4mm}
    \item $f_n$ is the normal force applied in equal-and-opposite fashion to each contacting geometry.
    \item $d$ is the penetration depth between two contacting geometries.
    \item $w$ is the transition region width specified in the block.
    \item $d\prime$ is the first time derivative of the penetration depth.
    \item $k$ is the normal-force stiffness specified in the block.
    \item $b$ is the normal-force damping specified in the block.
    \item $s(d,w)$ is the smoothing function.
\end{itemize}
Friction forces are modeled using a \textit{Smooth Stick-Slip} model with coefficient of static friction of $0.7$, coefficient of dynamic friction of $0.5$ and critical velocity $1\times 10^{-3}~m/s$. The stick-slip friction force is calculated by:
$$
| f_{f} |=\mu\cdot| f_{n} |
$$
where,
\begin{itemize}
\addtolength\itemsep{-4mm}
    \item $f_f$ is the frictional force.
    \item $f_n$ is the normal force.
    \item $\mu$ is the effective coefficient of friction.
\end{itemize}
At high relative velocities, the value of the effective coefficient of friction is close to that of the coefficient of dynamic friction. At the critical velocity, the effective coefficient of friction achieves a maximum value that is equal to the coefficient of static friction. From this block we can output contact information such as contact signal, penetration and separation distances, normal and friction force, relative normal velocity and relative tangential velocity, and rotations and translations of the follower and base frames. For our implementation, we are only sensing contact signal, normal and friction force, relative tangential velocity and follower frame rotation and translation. The contact signal gives a binary output; $0$ if there's no contact or $1$ during contacts. The detected friction force is a $1\times N$ matrix which just gives us the magnitude of the force. We use the tangential velocity vector, which is a $2\times N$ matrix, to get the unit vector which can be multiplied to get the friction force components, which when concatenated with the normal force ($1\times N$ matrix) gives us a $3\times N$ matrix of contact forces between any two surfaces. This is then rotated to the ground frame to help us analyse contact data and visualise it better. We set this entire model as a library block which becomes easier to make uniform changes across all the simulation environments we have set up for different experiments. 
\begin{figure}
    \centering
    \includegraphics[width=1\linewidth]{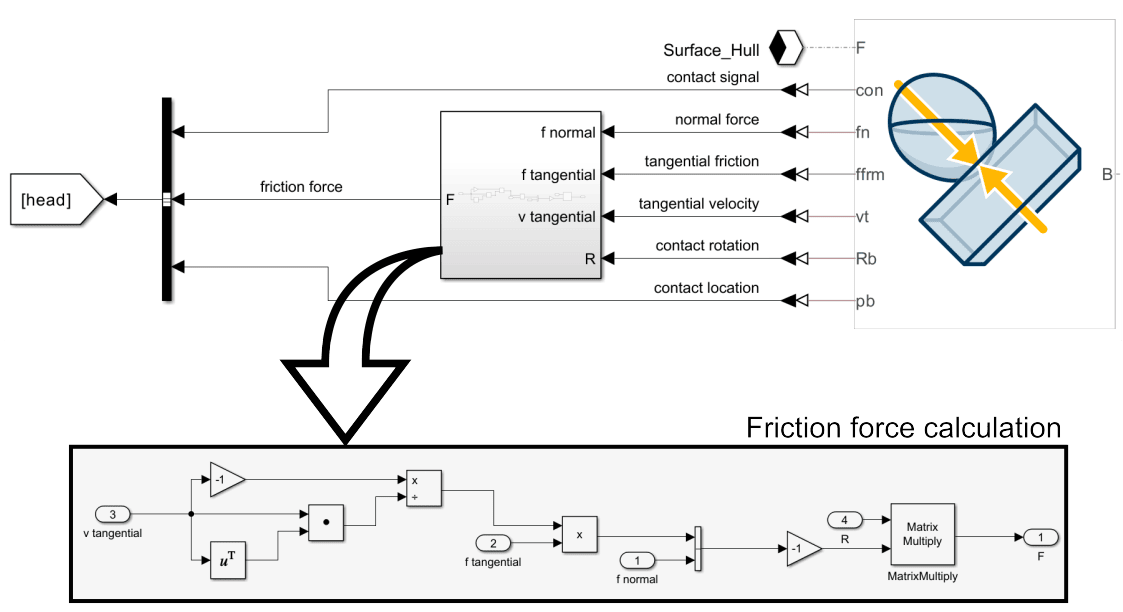}
    \caption{Simscape contact force block}
    \label{fig:simscape-contact}
\end{figure}

\subsection{Environment setup for Loco-manipulation}
\label{subsection:sim-environment}
Using the actuator model library block and the contact model library block, we can create simulation environments for our experimentation. We can directly connect the robot body hulls and the ground hull to the contact model block to model contacts between robot and ground and simultaneously output the contact data to Matlab workspace for further analysis. We have different output variables to store data for robot contacts with ground, box contacts with ground, robot contacts with box, robot joint data, robot pose data and box pose data. Considering the accuracy of Simscape model to match the actual robot movement, it becomes crucial to collect contact information and toqrue data from our simscape model since we don't have any IMUs or Force sensors on our hardware yet. We use the data we get from our Simscape model for contact model analysis. We can change the initial pose of the robot as per our requirements. In case of lateral rolling or sidewinding gaits we have the robot oriented in a way that the yawing joints are parallel to the ground plane. In case of our loco-manipulation gaits, we rotated the robot by $90~ ^{o}$. For setting up the environment for loco-manipulation, we have a model which contains the object, the platform and the ramp. The object being manipulated is modeled as a solid box of weight $0.5$ kg, with a replica of the tail module attached to the side of the box. The head module latches onto this module on the box while performing tasks like placing the box on the platform or picking the box from the platform. We have a similar setup to model contacts between the box and the ground. We also have a contact model which gives us information of contacts between the robot and the box which allows us to estimate which modules of COBRA are in contact with the box during locomotion for different gaits which gives us another metric to compare the gaits. The platform used to pick/place the box is placed at a height of $0.3~m$ and the ramp is inclined at an angle of $16.7^{~o}$ with maximum elevation of $0.6~m$. These both are modelled as solid objects with fixed positions with respect to the ground frame. We also have a separate contact model library block for the ramp with the robot and the box each. All these blocks are connected to an inertial frame (which is the ground frame in our case), a uniform gravity block with a constant acceleration of gravity of $9.8 ~m/s^2$ in the negative z axis, and a solver configuration block. 
\begin{figure}
    \centering
    \includegraphics[width=1\linewidth]{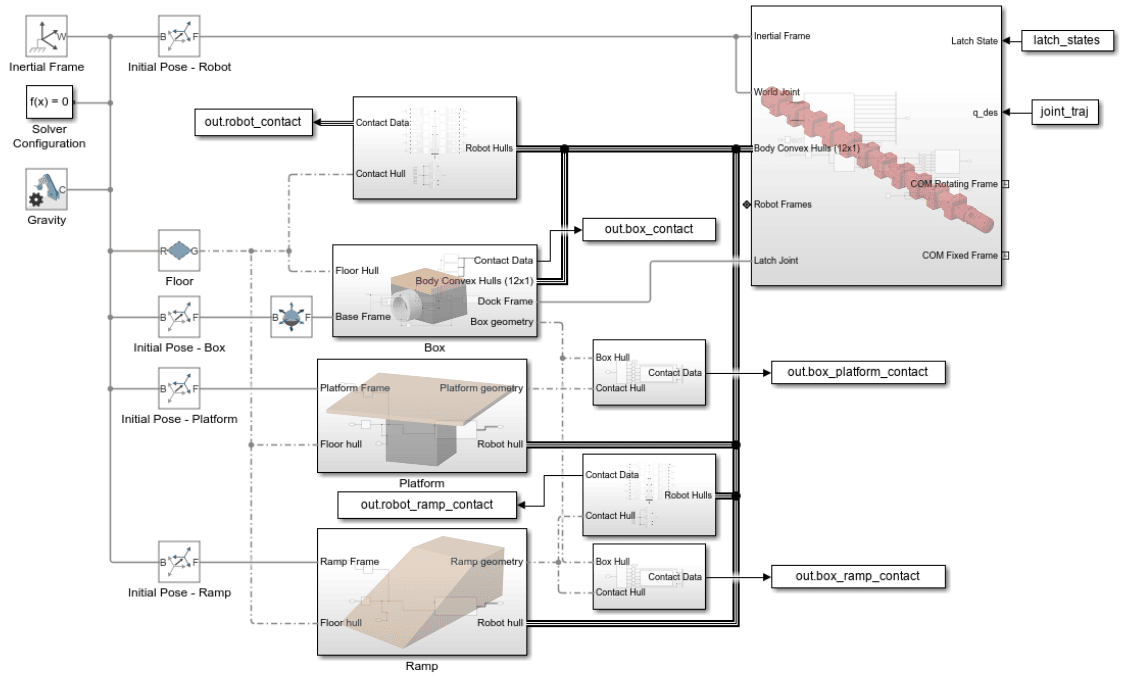}
    \caption{Simscape model setup}
    \label{fig:simscape-setup}
\end{figure}
Each physical network represented by a connected Simscape™ block diagram requires solver settings information for simulation. The Solver Configuration block specifies the solver parameters that your model needs before you can begin simulation. We have the Equation Formulation (specifies how solver treats sinusoidal variables) setup at Frequency and Time which makes the simulation faster, a consistency solver factor of $1\times 10^{-9}$ provides a scaling factor for determining how accurately the algebraic constraints are to be satisfied at the beginning of simulation and after every discrete event and by keeping the "Apply filtering at 1-D/3-D connections when needed" box checked, the solver automatically applies input filtering to the signal entering the Simulink-PS Converter block to obtain this additional derivative with a filtering time constant parameter of $0.001$ which provides the time constant for the delay. A solver applies a numerical method to solve the set of ordinary differential equations that represent the model. Through this computation, it determines the time of the next simulation step. In the process of solving this initial value problem, the solver also satisfies the accuracy requirements that you specify. An extensive set of fixed-step and variable-step continuous solvers are provided in Simscape, each of which implements a specific ODE solution method. The appropriate solver for simulating a model depends System dynamics, Solution stability, Computation speed, Solver robustness. By choosing the auto-solver, Simscape picks the best numerical method and a fixed timestep depending on the given model. In case the solver is not satisfactory, we have the freedom of choosing our own numerical integration method and timestep value in the solver configurations. For our model, we have auto-solver where the dynamics are solved using MATLAB's \textit{ode45} with a fixed timestep of $1 \times 10^{-4}$ seconds which are the default settings. We implemented a variable step solver for our model but it seemed to slow down the simulation. Even with the auto-solver setup, the total simulation time varies depending on the contacts, the more number of contacts we have defined in our model, the slower the simulation runs. We can speed up the simulation by increasing the timestep but it affects the accuracy of simulation which may not be a desirable tradeoff considering the data we use for analysis.
\begin{figure}
    \centering
    \includegraphics[width=1\linewidth]{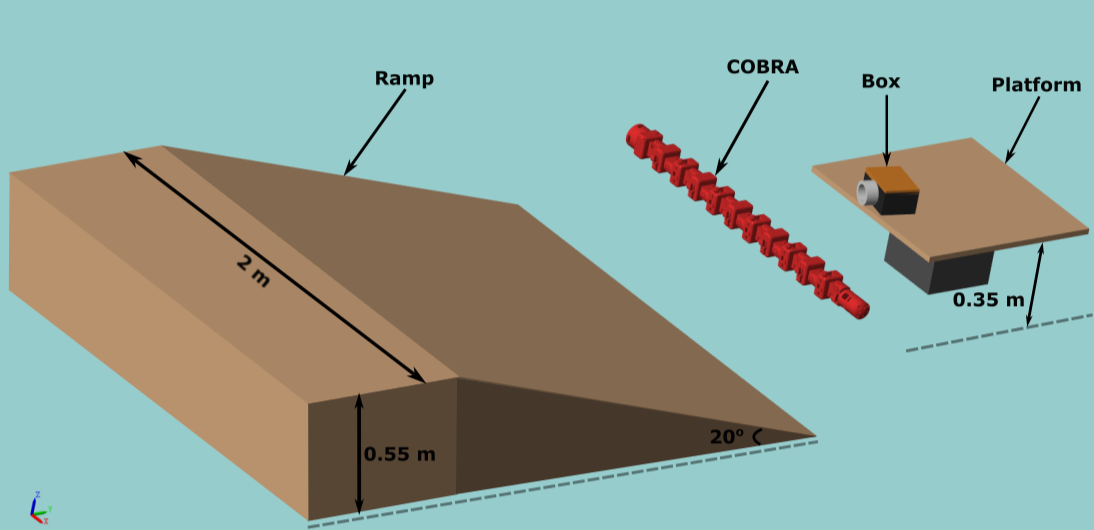}
    \caption{Simscape environment setup for loco-manipulation problem}
    \label{fig:simscape-locomanip}
\end{figure}
\section{Open-loop CPG gaits}
\label{susbection:CPG gaits}
\begin{figure}
    \centering
    \includegraphics[width=0.9\linewidth]{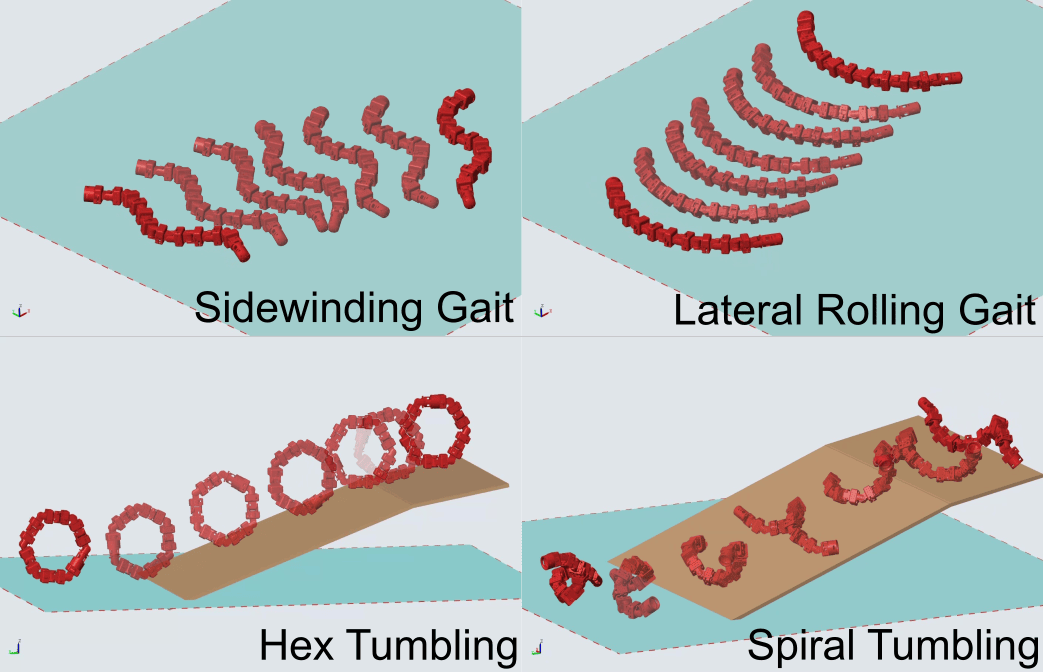}
    \caption{Various gaits implemented on COBRA}
    \label{fig:simscape-gaits}
\end{figure}

The COBRA hardware currently doesn't support a closed loop controller. In order to achieve desired movements we generate gaits on our COBRA model using an open-loop Central Pattern Generator (CPG). CPG produces a rhythmic output for a fixed given input. We use the same parameters in our simscape model and hardware. The simscape model gives a good approximation to the feasibility of a certain gait which can then be directly implemented on our hardware with a high accuracy. We generate gaits using a sinusoidal wave at each joint with fixed values for amplitude, frequency and phase angles. By varying these CPG parameters we obtain different gaits. After observing the sidewinding gaits in snakes, we determined that the same can be replicated by having one sinudsoidal wave which passes through the body in a plane parallel to the ground and another sinusoidal wave in a plane perpendicular to the ground and parallel to the direction of locomotion. The horizontal sine wave (in the plane parallel to ground) is the one that drives the robot to move forward whereas the sine wave in the vertical plane enables alternating anchor points. Implementing these simultaneously generates a gait which imitates sidewinding in snakes. For sidewinding (shown in Fig. \ref{fig:simscape-gaits}), we have an amplitude of $60^{~o}$ for yawing joints and $14^{~o}$ for pitching joints, frequency of $0.5$ and phase angles $\pi /2 * [0, 0, 1, 1, 2, 2, 3, 3, 0, 0, 1]$. In case of lateral rolling, all the joints in the same plane needed to move at the same time and the moment the robot flips by $90^{~o}$, the alternate joints should carry on the same movement. We have a sine wave that goes through all yawing joints simultaneously with no phase difference between them and have the same sine wave go through the pitching joints after one wave cycle, hence the yawing and pitching joints have a phase difference of $\pi /2$. The amplitude for this gait controls the shape of the lateral rolling which can be adjusting according to the environments that the COBRA operates in and the application for which we are using COBRA. For rolling gaits, we have the same amplitude of $20^{~o}$ for a C-shape lateral rolling (depicted in the Fig. \ref{fig:simscape-gaits}) for pitching and yawing joints, a phase difference of $\pi /2 * [0, 1, 0, 1, 0, 1, 0, 1, 0, 1, 0]$. We also have fixed angles with makes the COBRA go into the spiral configuration or the hex configuration which were determined by simple geometry (shown in Fig. \ref{fig:simscape-gaits}). These gaits are used for rolling down the slope without having to actively engage joints and conserving energy.


\chapter{Contact Modelling}
\label{chap:modelling}

\begin{figure}
    \centering
    \includegraphics[width=1\linewidth]{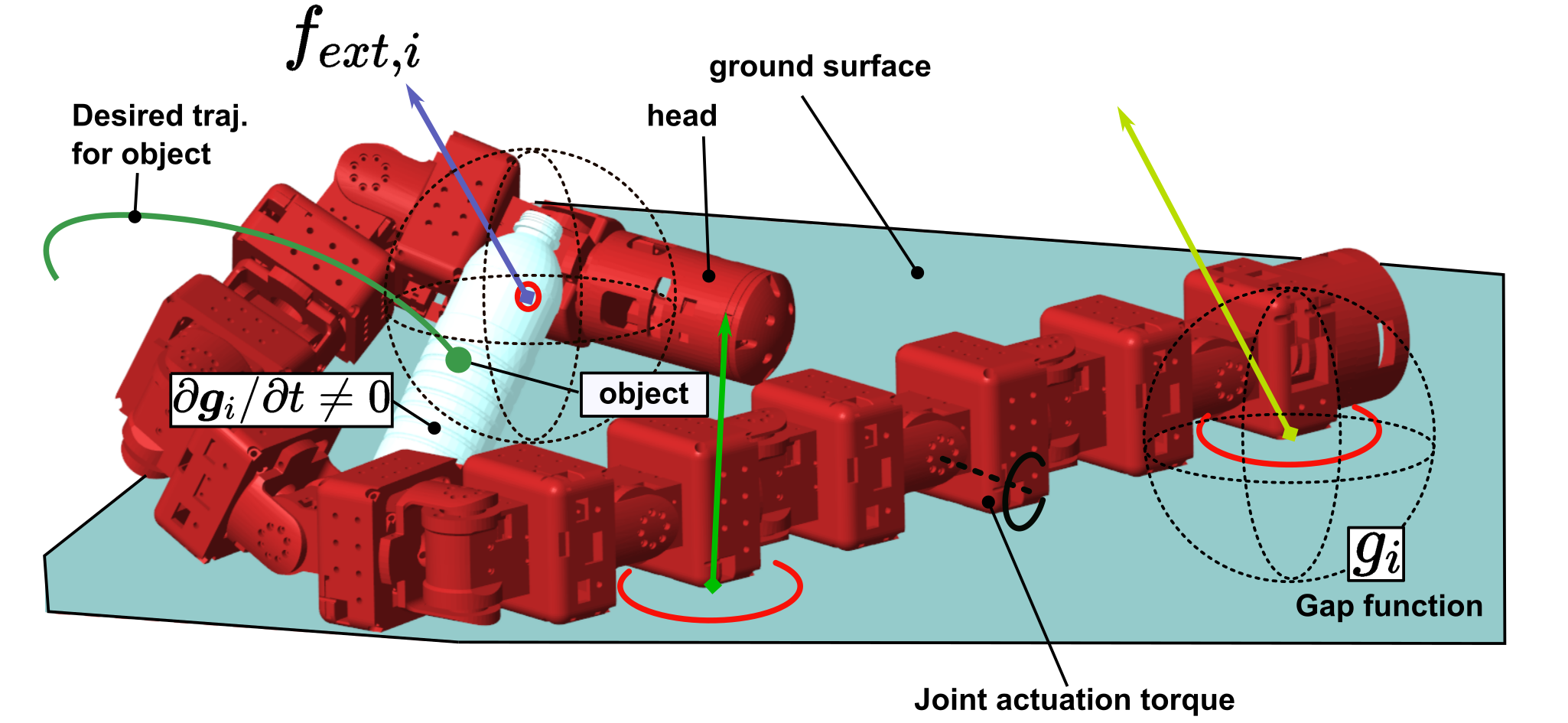}
    \caption{Full-dynamics model parameters in the object manipulation task considered in this paper}
    \label{fig:full-dyn}
\end{figure}
The motion dynamics governing the COBRA snake robot, equipped with 11 body joints, are succinctly captured in the following equations of motion \cite{salagame_how_2023}:
\begin{equation}
    \begin{aligned}
        \bm M(\bm q) \dot{\bm u} - \bm h(\bm q, \bm u, \bm \tau) &= \sum_i \bm J_i^{\top}(\bm q) \bm f_{ext,i},\\
        \bm h(\bm q, \bm u, \bm \tau) &= \bm C(\bm q,\bm u)\bm u + \bm G(\bm q) + \bm B(\bm q)\bm \tau
    \end{aligned}
    \label{eq:eom}
\end{equation}
In this expression, the mass-inertia matrix $\bm M$ operates in a space of dimensionality $\mathbb{R}^{17 \times 17}$, the terms encompassing centrifugal, Coriolis, gravity, and actuation ($\bm \tau$) are succinctly represented by $\bm h \in \mathbb{R}^{17}$, the external forces $\bm f_{ext,i}$ and their respective Jacobians $\bm J_i$ reside in the space $\mathbb{R}^{3 \times 17}$. 

In the object manipulation problem depicted in Fig.~\ref{fig:full-dyn}, the external forces stem solely from active unilateral constraints, such as contact forces between the ground surface and the robot or between a movable object and the robot. This assumption conveniently establishes a complementary relationship, where the product of two variables, including force and displacement, in the presence of holonomic constraints is zero, between the separation $\bm{g}_i$ (the gap between the body, terrain, and object) and the force exerted by a hard unilateral contact.

The concept of normal cone inclusion on the displacement, velocity, and acceleration levels from \cite{studer_numerics_2009} permits the expression:
\begin{equation}
    \begin{aligned}
        -\bm g_i & \in \partial \Psi_{i}\left(\bm f_{ext,i}\right) \equiv \mathcal{N}_{\mathcal{F}_i}\left(\bm f_{ext,i}\right) \\
        -\dot{\bm g}_i & \in \partial \Psi_{i}\left(\bm f_{ext,i}\right) \equiv \mathcal{N}_{\mathcal{F}_i}\left(\bm f_{ext,i}\right) \\
        -\ddot{\bm g}_i & \in \partial \Psi_{i}\left(\bm f_{ext,i}\right) \equiv \mathcal{N}_{\mathcal{F}_i}\left(\bm f_{ext,i}\right)
    \end{aligned}
    \label{eq:normal-cone-inclusion}
\end{equation}
where $\Psi_i(.)$ denotes the indicator function. The gap function $\bm{g}_i$ is defined such that its total time derivative yields the relative constraint velocity $\dot{\bm{g}}_i = \bm W_i^{\top} \bm u + \zeta_i$, where $\bm W_i = \bm W_i(\bm q, t) = \left( \partial \bm{g}_i / \partial \bm q \right)^{\top}$ and $\zeta_i = \zeta_i(\bm q, t) = \partial \bm{g}_i / \partial t$. 

In the context of the primary objectives of loco-manipulation with COBRA, various conditions of the normal cone inclusion as described in Eq.~\ref{eq:normal-cone-inclusion} were explored. In scenarios where nonimpulsive unilateral contact forces are employed to manipulate rigid objects (e.g., the box shown in Fig.~\ref{fig:full-dyn}), $\partial \bm{g}_i / \partial t \neq 0$. This factor holds significant importance in motion planning considered in this work and is enforced during optimization.

The total time derivative of the relative constraint velocity yields the relative constraint accelerations $\ddot{\bm g}_i=\bm W^{\top} \dot{\bm u}+\hat{\zeta}_i$ where $\hat{\zeta}_i=\hat{\zeta}_i(\bm q, \bm u, t)$. We describe a geometric constraint on the acceleration level such that the initial conditions are fulfilled on velocity and displacement levels:
\begin{equation}
    \begin{aligned}
        &\bm g_i(\bm q, t)=0, \\
        &\dot{\bm g}_i=\bm W_i^\top \bm u+\zeta_i=0, \\
        &\ddot{\bm g}_i=\bm W_i^\top \dot{\bm u}+\hat \zeta_i=0,\\
        &\dot{\bm g}_i\left(\bm q_0, \bm u_0, t_0\right)=0,\\
        &\partial \bm{g}_i / \partial t \neq 0
    \end{aligned}   
\end{equation}
which implies that the generalized constraint forces must be perpendicular to the manifolds $\bm g_i=0$, $\dot{\bm g_i}=0$, and $\ddot{\bm g}_i=0$. This formulation directly accommodates the integration of friction laws, which naturally pertain to velocity considerations. We divide the contact forces into normal and tangential components, denoted as $\bm f_{ext,i}=\left[\begin{array}{ll}f_{N,i}, & \bm f_{T,i}^{\top}\end{array}\right]^{\top} \in \mathcal{F}_{i}$. 

In this context, the force space $\mathcal{F}_i$ facilitates the specification of non-negative normal forces ($\mathbb{R}_{0}^+$) and tangential forces adhering to Coulomb friction $\left\{\bm f_{T,i} \in \mathbb{R}^2, ||\bm f_{T,i}||<\mu\left|f_{N,i}\right| \right\}$, with $\mu$ representing the friction coefficient.

The underlying rationale behind this approach is that while the force remains confined within the interior of its designated subspace, the contact velocity remains constrained to zero. Conversely, non-zero gap velocities only arise when the forces reach the boundary of their permissible set, indicating either a zero normal force or the maximum friction force opposing the direction of motion.

To proceed with the loco-manipulation problem considered here, it proves advantageous to reconfigure Eq.~\ref{eq:eom} into local contact coordinates (task space). This can be achieved by recognizing the relationship:
\begin{equation}
    \dot{\bm g}=\bm J_{\mathrm{c}} \bm u,     
\end{equation}
where $\bm g$ and $\bm J_c$ represent the stacked contact separations and Jacobians, respectively. By differentiating the above equation with respect to time and substituting Eq.~\ref{eq:eom}, we obtain:
\begin{equation}
    \ddot{\bm g} = \bm J_c \bm M^{-1} \bm J_c^{\top} \bm f_{ext} + \dot{\bm J}_c \bm u + \bm J_c \bm M^{-1} \bm h,
\end{equation} 
where $\bm G = \bm J_c \bm M^{-1} \bm J_c^{\top}$ -- the Delassus matrix -- signifies the apparent inverse inertia at the contact points, and $ \bm c = \dot{\bm J}_c \bm u + \bm J_c \bm M^{-1} \bm h$ encapsulates all terms independent of the stacked external forces $\bm f_{ext}$. At this point, the principle of least action asserts that the contact forces are determined by the solution of the constrained optimization problem:
\begin{equation}
    \begin{aligned}
        &\underset{\left\{\bm f_{ext,i},\bm u\right\}}{\operatorname{\textbf{minimize}}}~\frac{1}{2} \bm f_{ext}^\top \bm G \bm f_{ext}+\bm f_{ext}^\top \bm c \\
        &{\operatorname{\textbf{s.t.}}}\\
        &{\operatorname{\textbf{(1)}}~\bm M(\bm q) \dot{\bm u} - \bm h(\bm q, \bm u, \bm \tau) - \sum_i \bm J_i^{\top}(\bm q) \bm f_{ext,i}}=0\\
        &{\operatorname{\textbf{(2)}}}~\|\bm q\|\leq q_{max}\\
        &{\operatorname{\textbf{(3)}}}~\|\bm \tau\|\leq \tau_{max}\\
    \end{aligned}
\end{equation}
where $q_{max}$ and $\tau_{max}$ denote maximum joint movements and actuation torques, respectively. In the above optimization problem, (1), (2), and (3) enforce dynamics agreement, kinematics restrictions, and actuation saturations, respectively. 

Subsequently, a time-stepping methodology facilitates the integration of system dynamics across a time interval $\Delta t$ while internally addressing the resolution of contact forces. We employ the shooting method to find the optimal joint positions $\bm u_{ref}$ for minimizing $\frac{1}{2} \bm f_{ext}^\top \bm G \bm f_{ext}+\bm f_{ext}^\top \bm c$, ensuring that the generalized contact forces $\bm f_{ext}$ are orthogonal to gap functions and their derivatives.

 \chapter{Results}
\begin{figure*}
    \centering
    \includegraphics[width=1\linewidth]{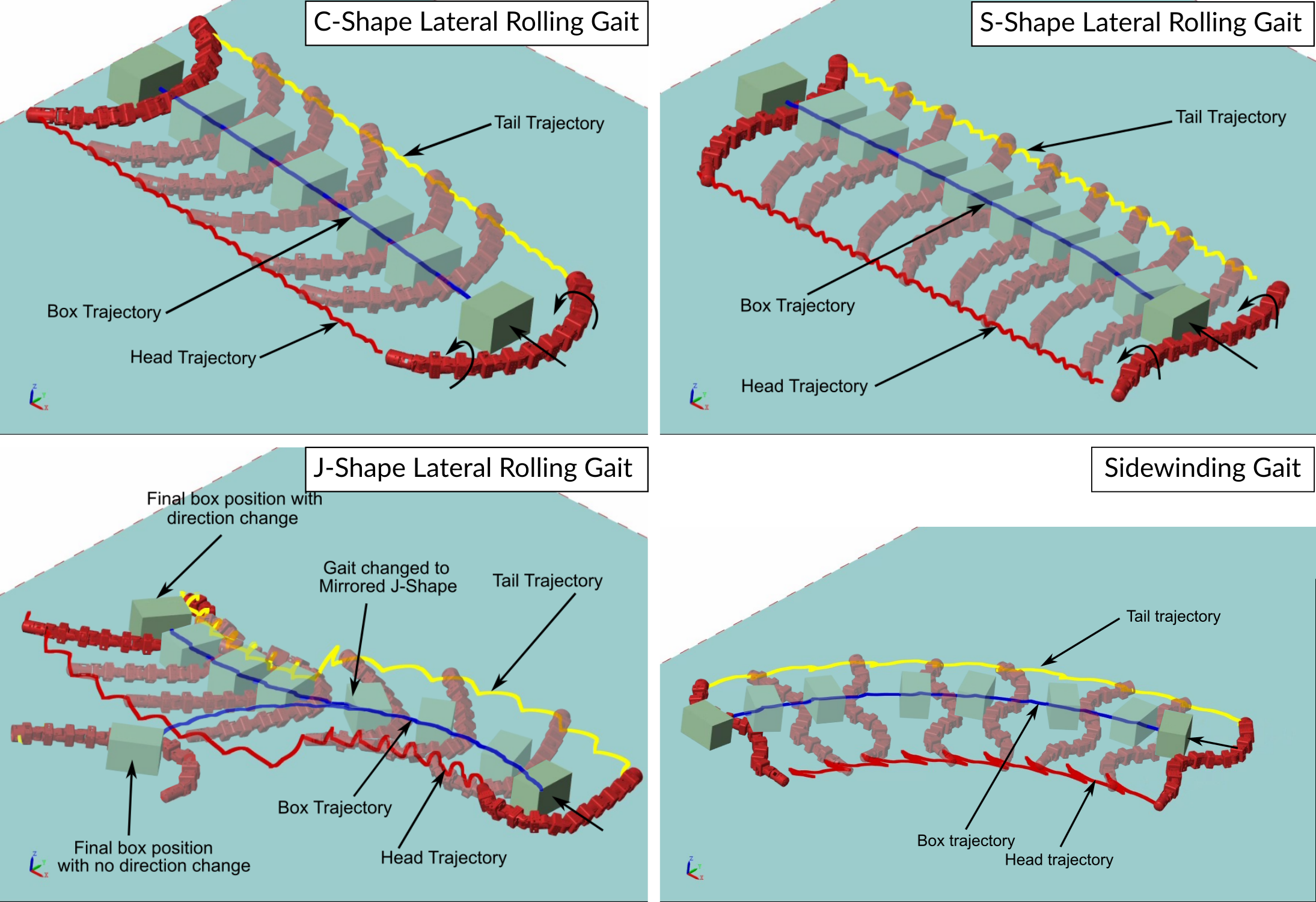}
    \caption{Snapshots depicting simulated forward box push utilizing various gaits executed in Matlab. The J-shape gait is an asymmetric variation of the C-gait, which allows changing the direction of movement of the box by mirroring the gait.}
    \label{fig:sim-snapshots}
\end{figure*}

\begin{figure*}
    \centering    \includegraphics[width=1\linewidth]{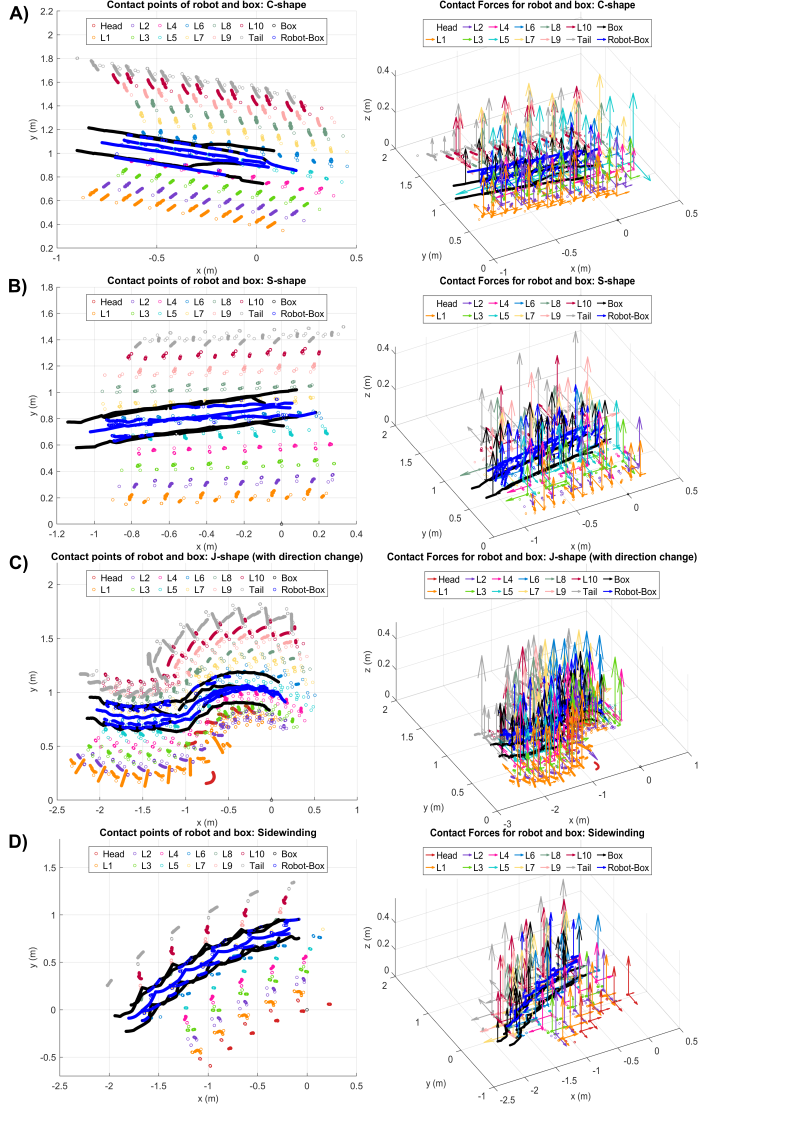}
    \caption{This image depicts the contact points and unilateral ground reaction forces during (A): C-shaped gaits, (B) S-Shaped Gait, (C) J-Shaped Gait, (D) Sidewinding gait, performed by the high-fidelity COBRA model simulated in the MATLAB environment. The contact forces consist of tangential forces ($\bm f_{T,i}$) and normal forces ($f_{N,i}$). L$x$ here refers to Link number $x$ on the robot, numbered from 1-10 starting from the head.}
    \label{fig:contacts}
\end{figure*}

We performed numerical dynamics integration of COBRA and a box interactions on flat ground. In Fig.~\ref{fig:sim-snapshots}, snapshots illustrating simulated forward box push using C-, S-, J-shaped lateral rolling, and sidewinding gait are presented. In these simulations, the goal is to move the box on the flat ground towards a specified point, and corresponding suitable joint commands are derived. The J-shaped lateral rolling gait is asymmetrical and can be mirrored to execute control on the direction in which the object is moved. The composite Fig.~\ref{fig:contacts} illustrates the contact points and unilateral ground reaction forces during (a) C-shaped gait, (b) S-shaped gait, (c) J-shaped gait, and (d) sidewinding gait, executed by the high-fidelity COBRA model simulated in the MATLAB environment. The contact forces encompass tangential forces ($\bm f_{T,i}$) and normal forces ($f_{N,i}$), where $Lx$ denotes Link number $x$ on the robot, numbered from 1 to 10 starting from the head. 

\begin{figure}
    \centering    \includegraphics[width=1\linewidth]{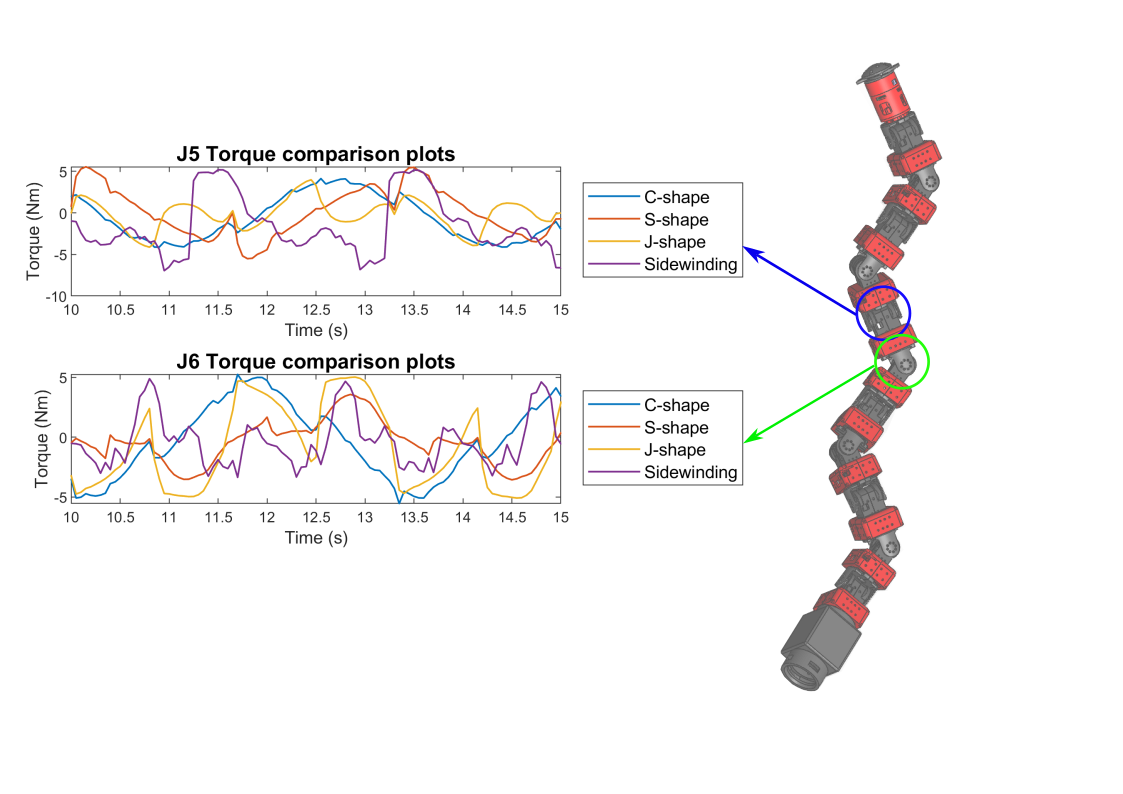}
    \caption{Depicts the torque profile for the central yawing (J6) and pitching (J5) joints for each gait. Other joints show similar profiles offset by a phase angle based on the executed gait.}
    \label{fig:Torque plots}
\end{figure}

Fig \ref{fig:Torque plots} shows the torques for the central yawing and pitching joints. The torque plots look identical across the robot for the yawing and pitching joints each with a phase difference. We get these values from our Simscape actuator model, since we do not have the capability of achieving this from hardware yet. Using these torques we can further analyse the efficiency of each of these rolling gaits.

\begin{figure}
    \centering    \includegraphics[width=0.5\linewidth]{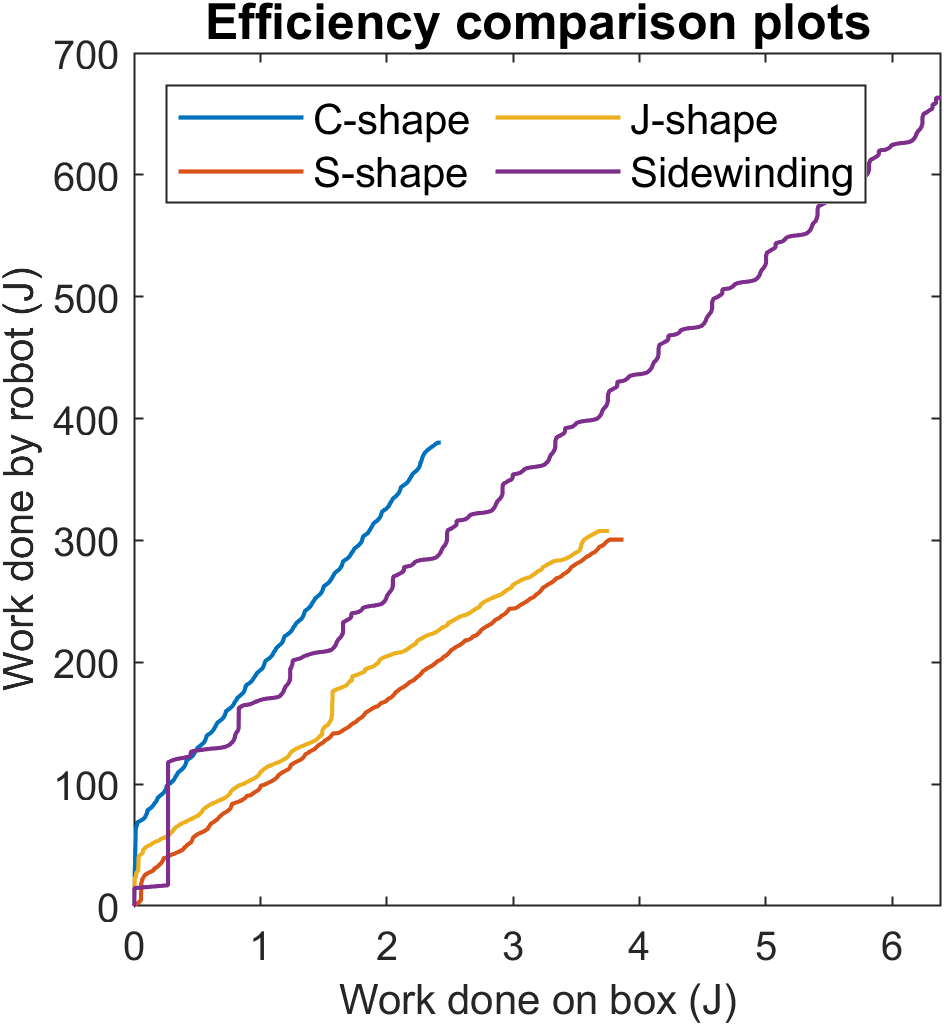}
    \caption{Depicts the efficiency of each gait based on the total work done by the robot for locomotion against total work done on the box. Larger slope indicates a less efficient gait as more work is done on locomotion in return for smaller work done on the box.}
    \label{fig:Work efficiency plots}
\end{figure}

\begin{figure}
    \centering    
    \includegraphics[width=0.5\linewidth]{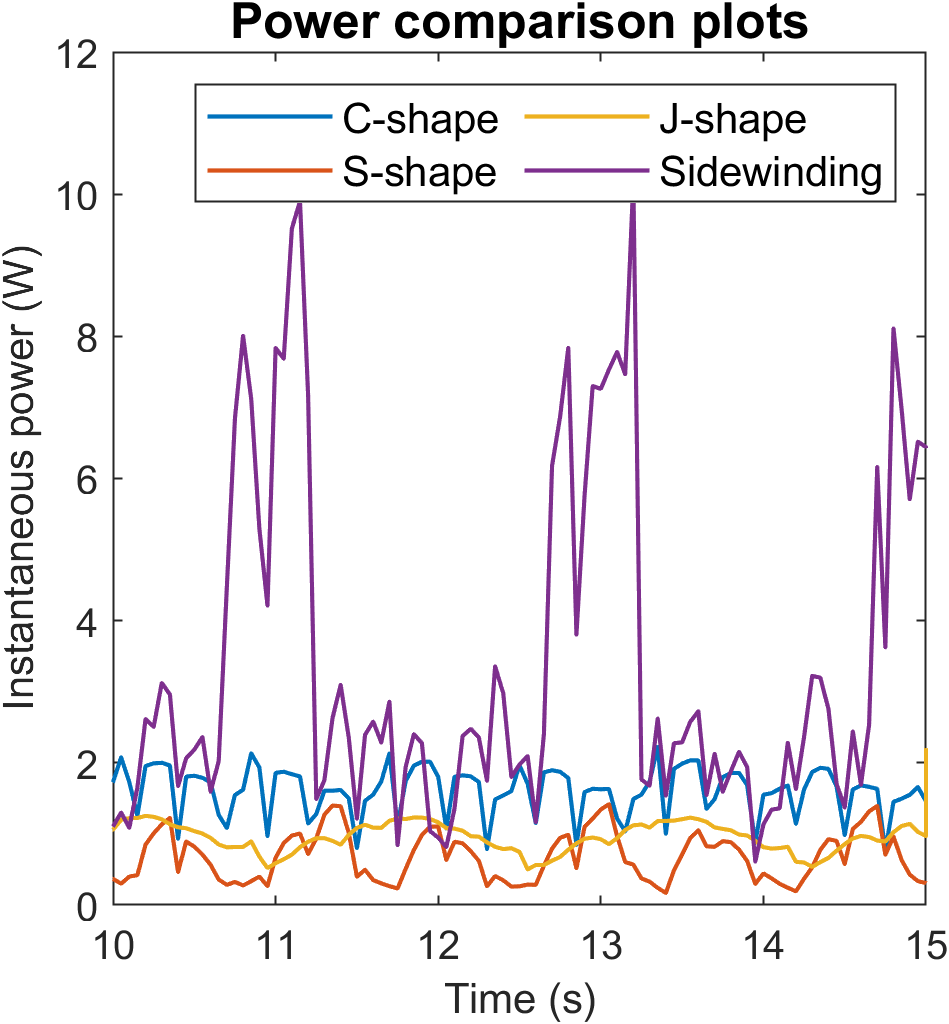}
    \caption{Shows the instantaneous power consumption by the robot to execute locomotion for each of the considered gaits.}
    \label{fig:Instantaneous power plots}
\end{figure}

Figures \ref{fig:Work efficiency plots} and \ref{fig:Instantaneous power plots} compare the efficiency of the four gaits in performing the prescribed task of moving the box. Figure \ref{fig:Work efficiency plots} plots the work done internally by the robot for locomotion against the work done by the robot on the box. The joint torques $\bm \tau$ for the gaits are shown in Fig. \ref{fig:Torque plots}. 

\begin{figure}
    \centering    \includegraphics[width=0.5\linewidth]{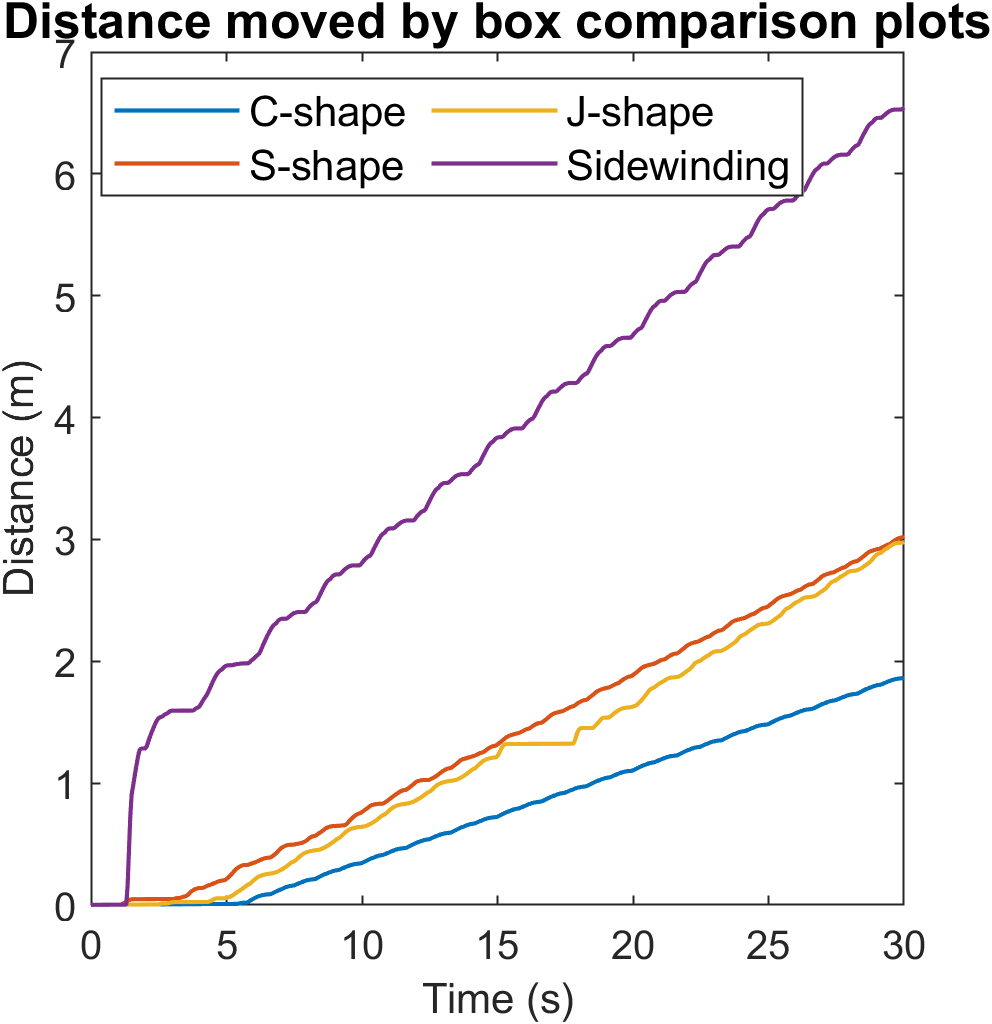}
    \caption{Depicts the total distance the robot moves the box for each gait in the same amount of time.}
    \label{fig:Distance comparison plots}
\end{figure}

More efficient gaits would need less work in locomotion to do more work on the box. In this respect, the S and J shape lateral rolling gaits perform more efficiently than other gaits, with Sidewinding doing the most work on the box, but consuming the most energy to do it. This is corroborated by the plot of instantaneous power for the robot for locomotion in Fig.~\ref{fig:Instantaneous power plots} that show the large peaks in instantaneous power from sidewinding, as compared to the more steady power consumption by other gaits. This results in a slower but more energy efficient loco-manipulation. Figure \ref{fig:Distance comparison plots} shows the relative distance moved by the box as a result of each gait operating for the same amount of time. 

\begin{figure}
    \centering    \includegraphics[width=1\linewidth]{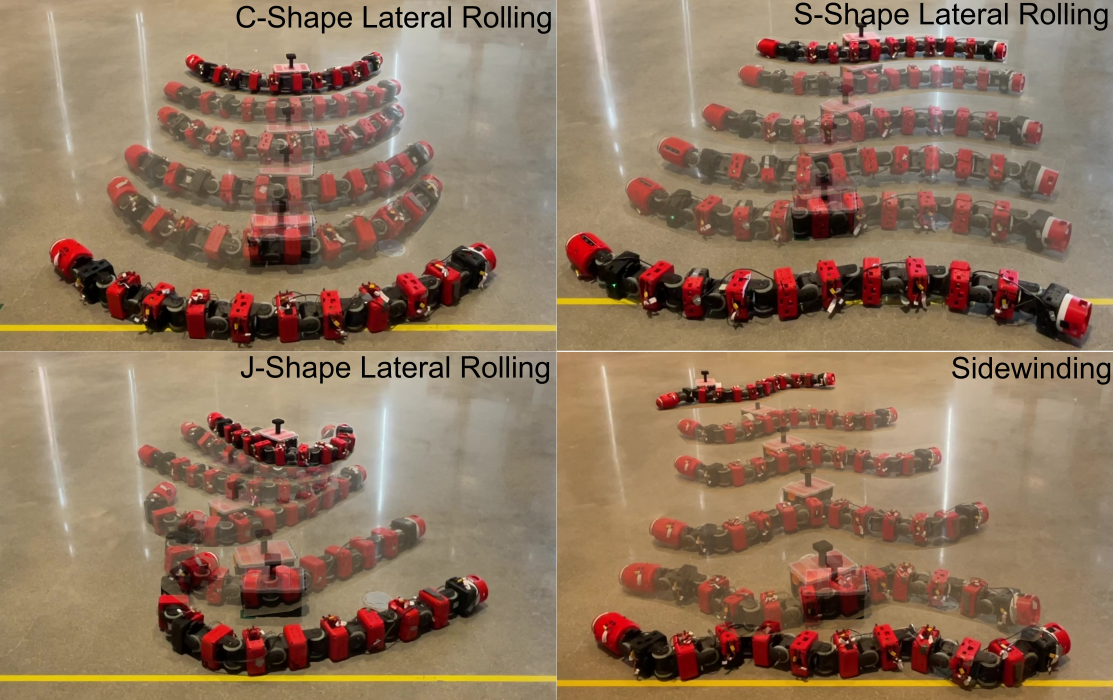}
    \caption{Shows the hardware implementation of the lateral rolling gaits performed in Simscape model}
    \label{fig:hardware}
\end{figure}

\begin{figure}
    \centering    \includegraphics[width=1\linewidth]{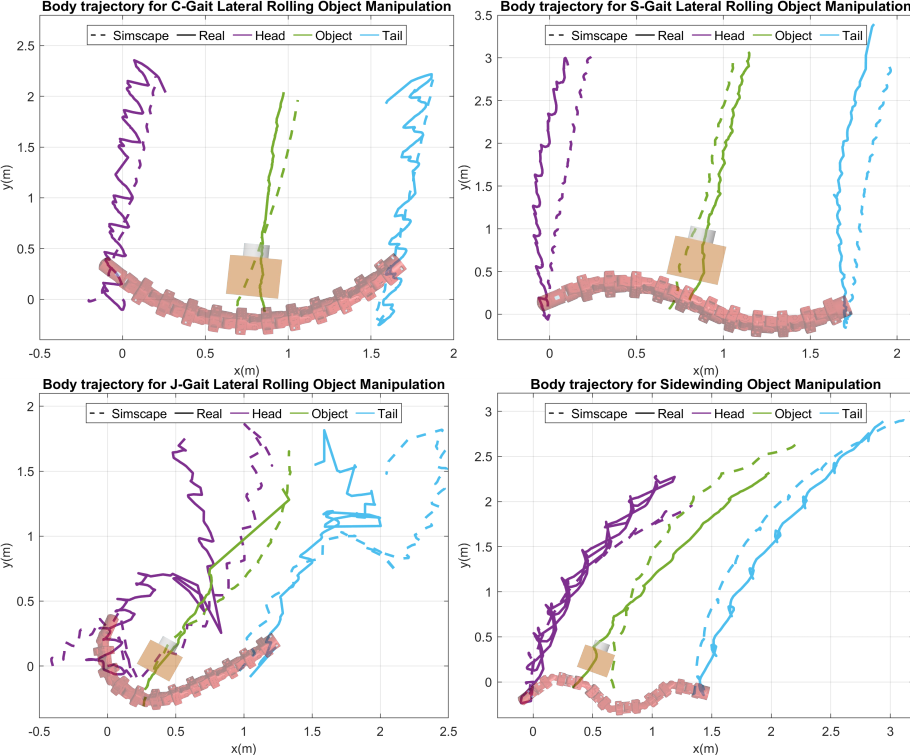}
    \caption{Shows comparison between simscape and real robot of robot and box for the various lateral rolling gaits on flat ground shown from top view}
    \label{fig:sim-real}
\end{figure}

We implemented the same gaits on hardware, as shown in fig \ref{fig:hardware}. We have our custom-made Optitrack attachments which gives us data for the head, tail and the box locations for our entire gait. We use this inform to make comparison between the head, tail and box trajectories from simscape and the same trajectories obtained from optitrack, which can be seen in fig \ref{fig:sim-real}. From this graphs we can conclude that we can accurately replicate the results from our simscape model on our hardware.

Once we achieved accurate hardware results, we demonstrate the object manipulation with COBRA by implementing the obtained desired trajectories in an open-loop fashion to show the effectiveness of our approach in preparation for the hardware implementation of the presented optimization model in closed-loop fashion. 

\begin{figure*}
    \centering
    \includegraphics[width=1\linewidth]{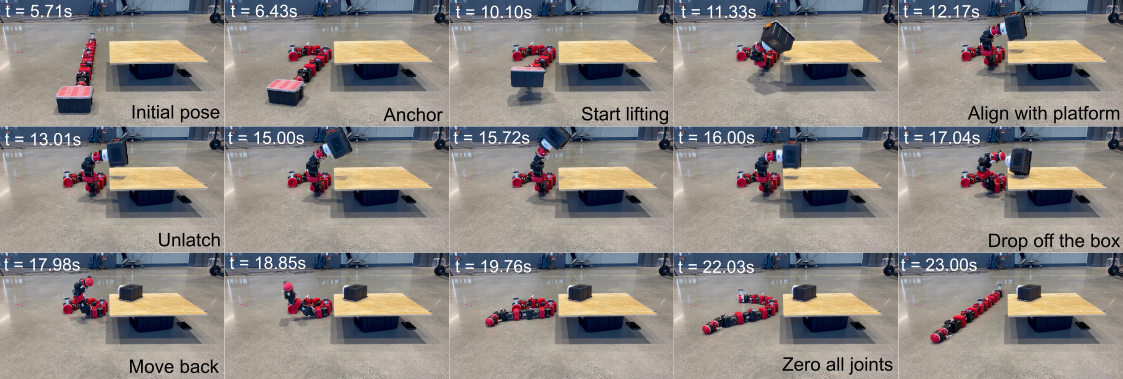}
    \caption{Snapshots of COBRA lifting a box and placing it on a raised platform.}
    \label{fig:liftnplace}
\end{figure*}

\begin{figure}
    \centering
    \includegraphics[width=1\linewidth]{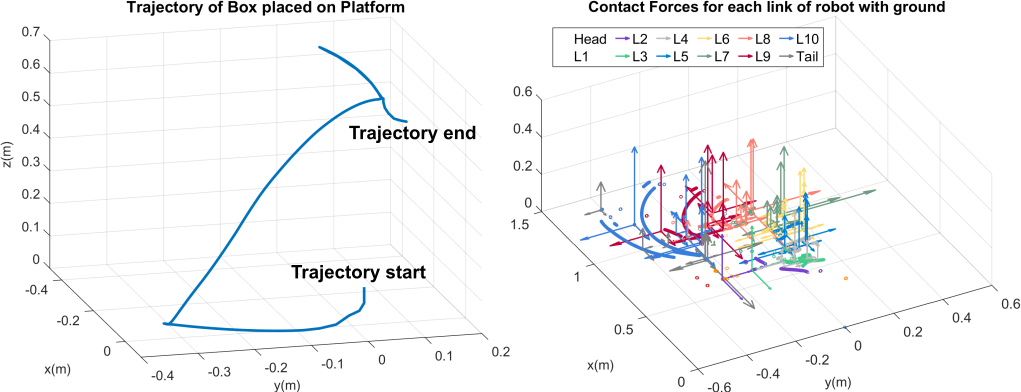}
    \caption{(Left) Shows the trajectory of the box while being lifted from ground and placed on a raised platform. (Right) Shows the contact forces between the robot and ground during the entire trajectory}
    \label{fig:platform-graphs}
\end{figure}

Figure~\ref{fig:liftnplace} shows COBRA lifting an object from the ground and placing it on a raised platform of height $0.3$m. The robot starts with the object docked to the head module using the docking module shown in Fig. \ref{fig:docking module}. It curls the tail to increase the region of support, allowing it to stay balanced while lifting the box up to a height of $0.4$m from the ground. Once it has moved the box over the platform, it unlatches from the object and shakes its head to dislodge it and deposit it on the platform. Fig. \ref{fig:platform-graphs} shows the box trajectory and robot contact forces for the entire trajectory.

\begin{figure*}
    \centering
    \includegraphics[width=1\linewidth]{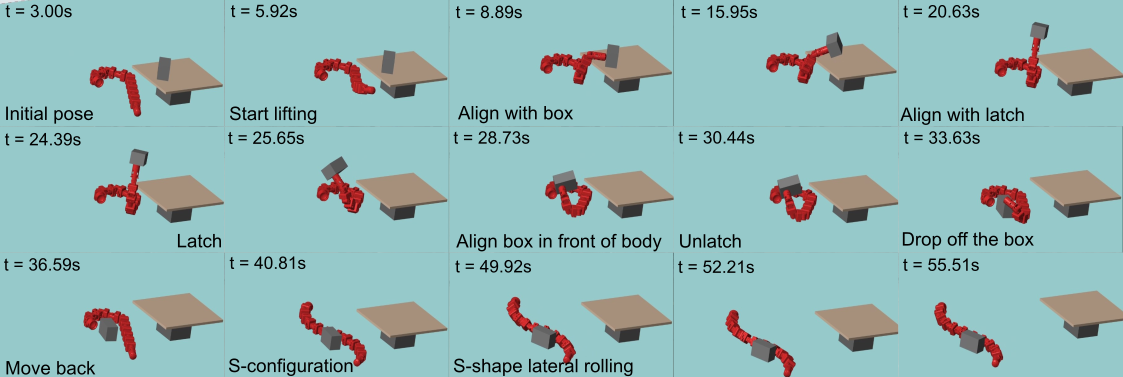}
    \caption{Snapshots depict simulation of COBRA lifting a box from a raised platform, placing it on the flat ground, and translating the box to a new location through continuous body-object interactions during slithering motions.}
    \label{fig:picknplace-sim}
\end{figure*}

\begin{figure*}
    \centering    \includegraphics[width=1\linewidth]{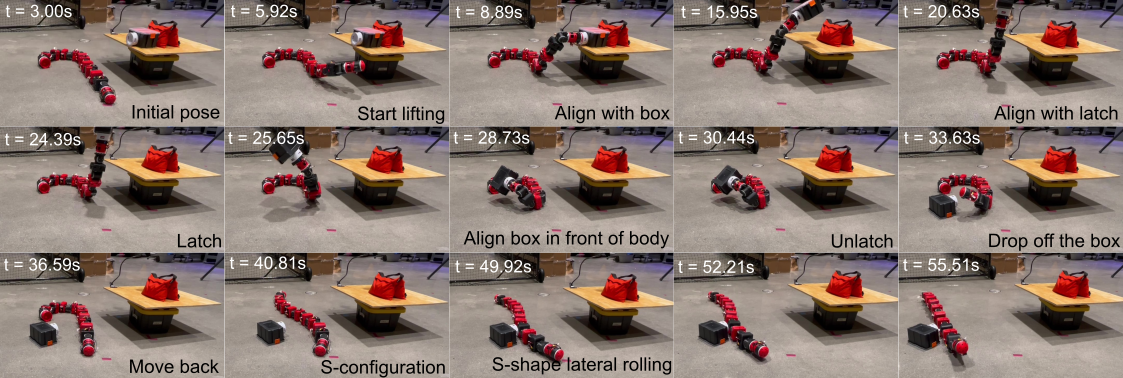}
    \caption{Snapshots depict COBRA lifting a box from a raised platform, placing it on the flat ground, and translating the box to a new location through continuous body-object interactions during slithering motions.}
    \label{fig:picknplace}
\end{figure*}

In Fig.~\ref{fig:picknplace-sim} and Fig.~\ref{fig:picknplace}, first in simulation, then in experiment, the robot starts on the ground with the object on the raised platform. The robot is able to, once again, curl its tail and lift the head to align itself with the docking module on the object. With some manipulation to allow the box to align to the latch, the robot can latch onto the box and place it in front of the body. From this position, COBRA can use its lateral rolling gait to move the object to a desired location. 

\begin{figure*}
    \centering
    \includegraphics[width=1\linewidth]{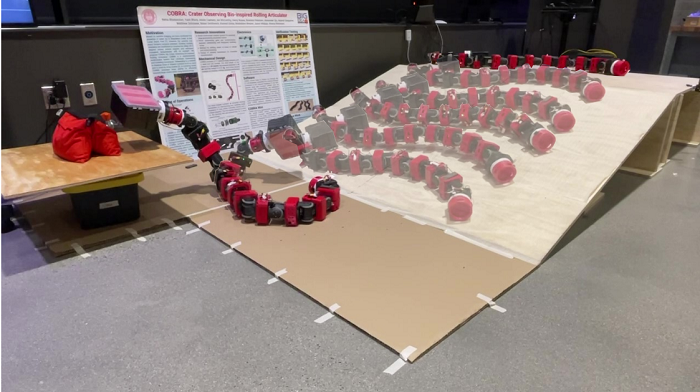}
    \caption{Snapshots capture COBRA lifting a box from a raised platform, setting it down on the ground, and ascending a ramp by continually pushing the box while lateral rolling.}
    \label{fig:ramp}
\end{figure*}

\begin{figure}
    \centering
    \includegraphics[width=1\linewidth]{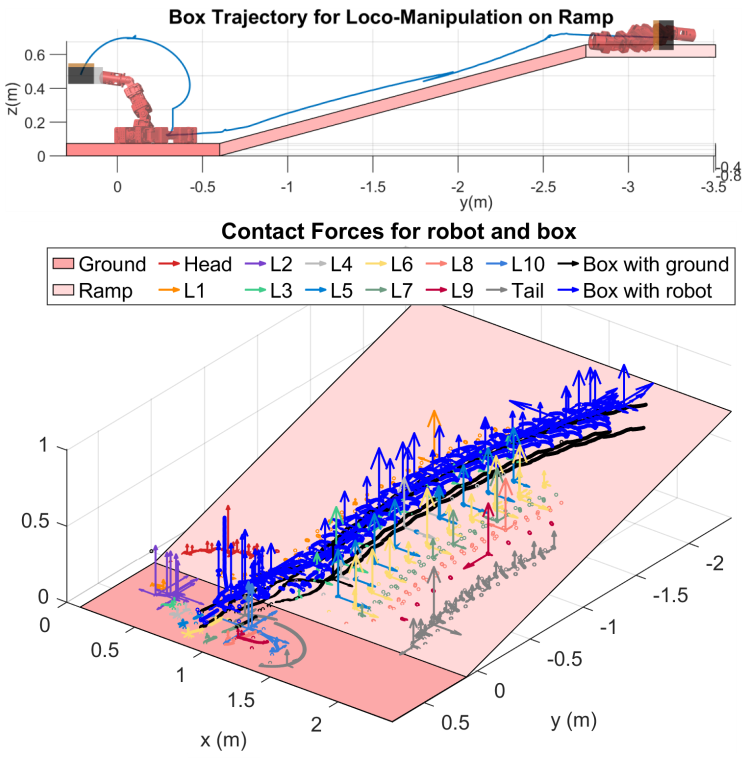}
    \caption{(Top) Shows the box trajectory during picking from platform, placing on ground and manipulating to the top of the ramp. (Bottom) Contact forces for the robot and box during picking the box from a raised platform, placing on the ground, and manipulation using s-lateral rolling gait up a ramp.}
    \label{fig:ramp-graphs}
\end{figure}

Fig.~\ref{fig:ramp} shows COBRA picking up the box from a raised platform, placing it in front of the body and moving the box to the top of a ramp. Fig.~\ref{fig:ramp-graphs} shows the box trajectory and contact forces for loco-manipulation up an inclined ramp.

\chapter{Conclusion}
Within the scope of this thesis, I have undertaken the development of finely-tuned open-loop gaits tailored specifically for achieving loco-manipulation capabilities on the COBRA platform. With its 11 actuated joints and integrated sensing and computation capabilities, COBRA emerges as a formidable platform, facilitating meticulous modeling for the manipulation of objects across flat surfaces. I worked on developing various gaits, each meticulously crafted to realize planar loco-manipulation objectives. Comparative analysis of these gaits, using metrics such as efficiency and instantaneous power plots, served as the cornerstone for determining the optimal gait patterns. Subsequently, I delved into the realm of more complex manipulation tasks, including the lifting a box from ground-level and placing it  onto elevated platforms. Having successfully achieved proficiency in these tasks, I embarked upon the integration of these skills with planar loco-manipulation techniques. Leveraging the capabilities of COBRA, I demonstrated maneuvers wherein the robot adeptly lifted a box from a raised platform, positioned it in front of itself, and then navigated to a predetermined location on level ground. Additionally, I replicated the same level of loco-manipulation proficiency when navigating up an inclined ramp. Furthermore, collaboration with Professor Alireza Ramezani and Adarsh Salagame, yielded significant strides in the development of a contact-rich optimization-based control methodology. This innovative approach holds promise for the creation of a closed-loop controller, offering enhanced precision and adaptability in executing manipulation tasks. In summation, this thesis encapsulates a multifaceted journey encompassing the refinement and application of locomotive manipulation techniques on the COBRA platform. The culmination of these efforts not only advances our understanding of robotic manipulation but also lays the groundwork for future developments in closed-loop control strategies.
\section{Challenges in existing setup}
\label{section:challenges}
Following our work on achieving loco-manipulation tasks with COBRA, I encountered several immediate challenges. Firstly, our current Simulink setup lacks a friction model that accurately mirrors real-world friction forces. Developing a controller that incorporates such a precise friction model is imperative to address this issue. Additionally, we observed compliance in joints due to 3D printed components and servo backlash. Compensation for servo backlash can be achieved through encoder data. However, we currently lack an estimator for 3D print compliance. Furthermore, with our current open-loop gaits, we face limitations in planning locomotion for the snake to accomplish tasks for which a gait hasn't been predefined. Lastly, the absence of any perception capabilities on the robot necessitates the predefinition of the locations for the box, platform, and ramp to design our open-loop gaits. 
\section{Future Scope}
Examining the challenges outlined in the previous section, there are viable strategies to tackle them. Firstly, leveraging our established mathematical model, we can proceed to develop a closed-loop controller. Within this controller framework, we can incorporate an estimator designed to accommodate joint compliance, thus enhancing overall control precision. Moreover, an avenue for improvement lies in the integration of sensory technologies. This includes the utilization of IMUs (Inertial Measurement Units) and force or tactile sensors to augment our feedback capabilities. By integrating these sensors, we gain valuable insights into the robot's interactions with its environment, facilitating more nuanced control and maneuvering. Furthermore, the incorporation of a RealSense camera system presents an opportunity to enhance perception capabilities. The RealSense camera can provide depth maps and enable object tracking within the operational environment. This additional sensory input empowers the robot with a more comprehensive understanding of its surroundings, thereby facilitating online and real-time path planning and re-planning based on the dynamic feedback received from these sensors.

\label{chap:conclude}



\printbibliography


\printindex

\end{document}